\documentclass[journal]{IEEEtran}
\ifCLASSOPTIONcompsoc

  \usepackage[nocompress]{cite}
\else
  \usepackage{cite}
\fi

\usepackage{booktabs,xcolor,siunitx}

\usepackage{times}
\usepackage{epsfig}
\usepackage{graphicx}
\usepackage{amsmath}
\usepackage{amssymb}
\usepackage{xcolor}
\usepackage{bbm}

\usepackage{multirow}
\usepackage{mathtools}
\usepackage{breqn}
\usepackage{kantlipsum}
\usepackage{algorithm}
\usepackage{algpseudocode}
\usepackage{tabularx}
\usepackage{subfigure}
\usepackage{siunitx}
\usepackage{amsmath}

\usepackage{times}
\usepackage{epsfig}
\usepackage{graphicx}
\usepackage{amsmath}
\usepackage{amssymb}
\usepackage[algo2e]{algorithm2e}
\usepackage{algorithm}
\usepackage{algpseudocode}
\usepackage{multirow}
\usepackage{mathtools}
\usepackage{breqn}
\usepackage{kantlipsum}
\usepackage{algorithm}
\usepackage{algpseudocode}
\usepackage{tabularx}
\usepackage{subfigure}
\DeclareMathOperator*{\mymin}{argmin}
\DeclareMathAlphabet\mathbfcal{OMS}{cmsy}{b}{n}
\usepackage{relsize}
\usepackage[algo2e]{algorithm2e}
\algnewcommand\algorithmicforeach{\textbf{for each}}
\algdef{S}[FOR]{ForEach}[1]{\algorithmicforeach\ #1\ \algorithmicdo}
\newcolumntype{Y}{>{\centering\arraybackslash}X}
\newcommand\myeq{\mathrel{\overset{\makebox[0pt]{\mbox{\normalfont\tiny\sffamily def.}}}{=}}}
\newcolumntype{P}{>{\centering\arraybackslash}c}
\definecolor{brown}{RGB}{215,183,133}
\definecolor{silver}{RGB}{211,211,211}
\usepackage{bm}

\title{A Generic Self-Supervised Framework of Learning Invariant Discriminative Features}

\author{Foivos~Ntelemis,
        Yaochu~Jin, \emph{Fellow}, \emph{IEEE},
        Spencer~A.~Thomas
\thanks{This project is funded by an EPSRC industrial CASE award (number 17000013) and Department for Business, Energy and Industrial Strategy through the National Measurement System (122416). (\textit{Corresponding author: Yaochu Jin})}
\thanks{F. Ntelemis is with the Department of Computer Science, University of Surrey, Guildford, GU2 7XH, United Kingdom. (Email: f.ntelemis@surrey.ac.uk)}
\thanks{Y. Jin is with the Faculty of Technology, Bielefeld University, 33619 Bielefeld, Germany. He is also with the Department of Computer Science, University of Surrey, GU2 7XH, United Kingdom. (Email: yaochu.jin@uni-bielefeld.de)}
\thanks{S. A. Thomas is with the National Physical Laboratory, Teddington, TW11 0LW, United Kingdom. He is also with the Department of Computer Science, University of Surrey, GU2 7XH, United Kingdom. (Email: spencer.thomas@npl.co.uk)}
}

\begin{document}

\maketitle

\begin{abstract}

Self-supervised learning (SSL) has become a popular method for generating invariant representations without the need for human annotations. Nonetheless, the desired invariant representation is achieved by utilising prior online transformation functions on the input data. As a result, each SSL framework is customised for a particular data type, e.g., visual data, and further modifications are required if it is used for other dataset types. On the other hand, autoencoder (AE), which is a generic and widely applicable framework, mainly focuses on dimension reduction and is not suited for learning invariant representation. This paper proposes a generic SSL framework based on a constrained self-labelling assignment process that prevents degenerate solutions. Specifically, the prior transformation functions are replaced with a self-transformation mechanism, derived through an unsupervised training process of adversarial training, for imposing invariant representations. Via the self-transformation mechanism, pairs of augmented instances can be generated from the same input data. Finally, a training objective based on contrastive learning is designed by leveraging both the self-labelling assignment and the self-transformation mechanism. Despite the fact that the self-transformation process is very generic, the proposed training strategy outperforms a majority of state-of-the-art representation learning methods based on AE structures. To validate the performance of our method, we conduct experiments on four types of data, namely visual, audio, text, and mass spectrometry data, and compare them in terms of four quantitative metrics. Our comparison results \textcolor{black}{demonstrate that the proposed method is effective and robust in identifying patterns within the tested datasets}.

\end{abstract}


\section{Introduction}

\IEEEPARstart{U}{nsupervised} learning describes the process of recognising hidden patterns from a collection of data-points without the requirement of laborious human annotations. Various unsupervised tasks have been studied from different perspectives, \textcolor{black}{e.g., clustering algorithms \cite{BEZDEK1984191,DBLP:conf/iccv/ComaniciuM99,Heller2005BayesianHC,Williams1999AMA,10.1007/978-3-642-33718-5_31} that aim to find homogeneous sub-populations of the dataset, traditional reduction strategies \cite{10.1162/089976699300016728, 2009nmf} that focus on reducing the length of features necessary to describe each point of the dataset while maintaining as much information as possible, and visualisation methods \cite{vanDerMaaten2008,tenenbaum_global_2000, roweis_nonlinear_2000} that map the initial high-dimensional data onto a usually two-dimensional space while preserving the original spatial neighborhood.}
\par

Representation learning, which also belongs to the group of dimensional reduction strategies, can be understood as a process that automatically extracts certain types of semantic information (features) from data to accomplish particular machine learning tasks. This includes tasks that require the handling of high-dimensional data types, such as visual or audio, which demonstrate non-linear properties. Aiming to replace linear traditional reduction strategies, autoencoders (AEs) \cite{HintonSalakhutdinov2006b, 10.1145/1390156.1390294,sparse, journals/jmlr/VincentLLBM10} are proposed to identify essential hidden information and deal with non-linearity. Typically, an AE framework consists of two modules: 1) an encoder module that projects the inputted data onto a latent representation (also called embedding space); and 2) a decoder module that reconstructs the input by reading the latent representation. Benefiting from the autoencoder or later generative adversarial networks (GANs), generative approaches \cite{Kingma2014,Higgins2017betaVAELB, NIPS2014_5423, DBLP:conf/iclr/BrockDS19, DBLP:journals/corr/RadfordMC15, 4270182} infer the latent space by following certain probabilistic representations. Despite the fact that generative models demonstrate superiority in learning the underlying patterns, their ability to learn invariant representation is limited. Invariant representations mean that the learned representations (e.g., features) are insensitive to some transformations in the input, for instance, rotation, shift, and translation of the object to be recognised in image data \cite{LKhc2020ContrastiveRL}. \par

Recently, self-supervised learning (SSL), a subset of representation learning, has gained increased attention due to its capability of leveraging neural network models and mapping multi-dimensional data onto an invariant representation. The learning process of SSL varies depending on the task and the distribution of the input data, which can be grouped into two major categories: 1) contrastive methods \cite{pmlr-v119-chen20j,He_2020_CVPR,2018arXiv180703748V, DBLP:journals/ijcv/ZhaoXWWTL20, pmlr-v9-gutmann10a}, which learn invariant representations by pairwise comparison of features obtained by the model, with the goal \textcolor{black}{of maximising} the similarities between positive pairs (features derived from the same instances with transformation functions applied in advance) and pushing negative pairs apart (features of dissimilar augmented instances); 2) grouping methods \cite{asano2020self,caron2020unsupervised,caron2018deep}, which first transform the input data with a transformation function to ensure invariance, and then cluster the transformed data into a large number of pseudo-classes. Both categories demonstrate impressive performances and are applicable to, after some modifications, various datasets such as visual\cite{caron2020unsupervised,pmlr-v119-chen20j,He_2020_CVPR,2018arXiv180703748V, DBLP:journals/ijcv/ZhaoXWWTL20, pmlr-v9-gutmann10a,asano2020self,caron2018deep}, audio \cite{pmlr-v9-gutmann10a,schneider2019wav2vec,baevski2020vqwav2vec}, or word/text \cite{mikolov2013efficient,chi2021infoxlm}. Nevertheless, these implementations require \textcolor{black}{problem-specific transformations of the data beforehand}, e.g., in the case of visual datasets, colour jittering, rotations\textcolor{black}{,} horizontal-flipping, and grey-scale conversion. Hence, their applicability to non-visual datasets is limited if such transformations are not allowed. Additionally, each method is specialised to a particular data type and some particular neural network architectures, and a generic SSL method for invariant representation learning still lacks. \par

Our motivation is to develop a generic framework for learning invariant representation via a self-supervised learning strategy applicable to various data types\textcolor{black}{,} including vector/dense data. To this end, the proposed framework avoids using any priori specified transformation function and instead suggests a training objective inspired by previous visual pretext studies \cite{asano2020self,caron2020unsupervised} that can learn discriminative representation based on self-labelling assignment (grouping methods). Unlike most existing methods relying on visual transformation functions, we impose invariance in representation via an internal augmentation mechanism realised by an adversarial training process. An internal augmentation mechanism is defined as a process in which the model transforms the input data-points without previously specifying any external transformation functions. The architecture of the proposed method can be seen as an autoencoder model by replacing the decoder part with a small classification module. Specifically, \textcolor{black}{the} first two variations of the same input data are computed by augmenting it with the help of virtual adversarial training (VAT)\cite{8417973}. VAT is a flexible\textcolor{black}{,} unsupervised process of generating adversarial perturbations without the requirement of label information. Then, the regularised Sinkhorn-Knopp algorithm \cite{NIPS2013_af21d0c9} is employed to compute , respectively,  two target distributions and exchange them to train the model's parameters in a contrastive fashion. The Sinkhorn-Knopp algorithm \cite{NIPS2013_af21d0c9} is used to prevent degenerate solutions (all data-points are assigned to the same classifier output) and ensure an equal distribution of each probabilistic index at the output of the classifier. We also introduce a policy that adjusts the regularisation term in the Sinkhorn-Knopp algorithm \cite{NIPS2013_af21d0c9} to control the smoothness of the generated target distribution within a pre-defined level. By using the introduced policy, a possible trivial grouping effect in the embedding domain that splits the data into a number of equally divided clusters is prevented. Finally, the framework learns the representation by predicting the same target distribution invariant to the augmented inputs in the presence of VAT perturbations. The full training process is carried out in an unsupervised end-to-end fashion. \par

As presented in the section on experimental result\textcolor{black}{s}, our proposed model is able to perform invariant representation learning for various data types, including visual, audio, text content, and mass spectrometry imaging data (MSI). Without relying on any ad hoc a priori transformations or a specific network architecture, our framework is capable of extracting meaningful discriminative features that outperform autoencoder structures. To validate the extracted features, the framework is compared with a range of generic methods and validated according to four quantitative measurements. \par

The main contributions of this work are summarised as follows:

\begin{enumerate}
    \item We \textcolor{black}{propose} a generic SSL framework that learns \textcolor{black}{unsupervised} invariant representation without using a transformation function dependent on specific data types. \textcolor{black}{Although the learning process is unsupervised, the proposed framework is capable of generating meaningful and highly separable features without the use of any human annotation. Additionally, in contrast to the previous studies, the proposed learning mechanism can be adapted to different neural architectures, such as convolutional neural networks (CNNs) and multi-layer perceptrons (MLPs).}
    
    \item \textcolor{black}{We replace the specific transformation functions for particular data types with a generic self-transformation process named virtual adversarial training (VAT) and include it in the proposed self-supervised framework as a flexible self-transformation mechanism without semantically modifying the input data. The introduced mechanism demonstrates the possibility to replace, when necessary, the prior transformation/augmentation functions.}
    \item \textcolor{black}{The} unsupervised adversarial training process is modified and proposed as an effective contrastive adversarial learning objective. This \textcolor{black}{enables the framework to learn} invariant representation through multiple augmented instances.
    \item We propose an \textcolor{black}{effective} adaptation policy to maintain the smoothness of the target distribution at a defined level by tuning the regularisation hyperparameter \textcolor{black}{of} the Sinkhorn-Knopp algorithm. This prevents the framework from sub-grouping the input data in the latent space due to the classification module embedded in the framework.
\end{enumerate}

\section{Related Work}

\subsection{Representation Learning}
\subsubsection{Generative models} Early studies propose the deep AE framework\cite{HintonSalakhutdinov2006b,10.1162/neco.2006.18.7.1527} \textcolor{black}{for learning meaningful features, an unsupervised learning strategy that} projects the initial input data onto a bottleneck embedding space through an encoder module and afterwards reconstructs the original input via the decoder module. Following the success of the AE frameworks, later studies extend the training process by including additional penalties either to regularise the sparsity of the embedding layer \cite{sparse,Olshausen1996,NIPS2006_2d71b2ae} or \textcolor{black}{to} reconstruct the augmented, by random perturbation, input points to the initial input \cite{10.1145/1390156.1390294,pmlr-v32-bengio14}. A second group of studies, known as generative models \cite{Kingma2014,Higgins2017betaVAELB, tolstikhin2018wasserstein,makhzani2016adversarial}, leverage the AE's architecture and regularise the latent space to approximate a probabilistic behaviour (e.g., a Gaussian distribution). In the family of generative models, the GAN frameworks \cite{NIPS2014_5423, DBLP:conf/iclr/BrockDS19,DBLP:journals/corr/RadfordMC15,DBLP:conf/iclr/DonahueKD17} are also composed of two modules, one generator and one discriminator. The generator generates \textit{fake} input data  with the aim of approximating the original distribution of the source inputs. By contrast, the discriminator is used to distinguish whether the output of the generator is from the true input data or from the fake input. As a result, the generator is trained to produce instances identical to the original data. Despite the success of the generative frameworks, their capability \textcolor{black}{for} learning invariant representation is still limited. Additionally, only one of the two modules is actually used as a feature extraction pipeline, and the other one is discarded after the completion of the training. \par

\subsubsection{Self-Supervised Learning (SSL)} In contrast to AEs and generative models, SSL only requires the implementation of an encoder. The early studies were based on the design of handcrafted pretext tasks, where the encoder is trained either to treat each sample as a unique class \cite{DBLP:journals/pami/DosovitskiyFSRB16}; to detect the rotation of the image \cite{DBLP:conf/iclr/GidarisSK18}; or to identify the solution of a jigsaw puzzle \cite{10.1007/978-3-319-46466-4_5}. Oord {\em  et al.} \cite{2018arXiv180703748V} introduced the \textit{noise contrastive estimation} (infoNCE) and the encoder is optimised by predicting the next patch of a series (image or audio) through an auto-regressive module. Due to the promising results, the infoNCE objective function has attracted a lot of attention \cite{caron2020unsupervised,pmlr-v119-chen20j,He_2020_CVPR,2018arXiv180703748V, DBLP:journals/ijcv/ZhaoXWWTL20,hu2021adco} to maximise the similarities between positive pairs of representations and minimise the similarities between negative pairs of representations within a Siamese style network. Here, a positive pair is defined as representations obtained from the same input with the prior application of a group of transformation operations, while a negative pair is defined as any pair drawn from samples with different indexes. In contrast to the infoNCE objective that requires a large number of negative pairs to avoid model collapse, later studies either proposed an objective function based on the Siamese triplet loss \cite{Wang2021Solving_iccv}, which requires fewer negative pairs, or remedied the above issue by using an additional prediction module on top of the comparing representation, stopping the computation of the gradient in one of the two Siamese architectures, or/and by using a momentum encoder \cite{grill2020bootstrap} \cite{grill2020bootstrap,chen2020exploring,tian2021understanding,pmlr-v139-ermolov21a}, or using a statistical whitening process \cite{pmlr-v139-ermolov21a}, to enhance the performance and prevent model collapse. As an alternative to the comparison of positive representations, the grouping training strategies learn representations by partitioning the given dataset into a number of pre-defined pseudo-classes. The grouping is accomplished either through the use of an optimal transport plan \cite{asano2020self,caron2020unsupervised}, e.g., using the Sinkhorn-Knopp algorithm \cite{NIPS2013_af21d0c9}, through the implementation of a \textit{k}-means algorithm, or by leveraging a training process through a self-supervised classification objective function \cite{amrani2021selfsupervised}. It should be pointed out that the aforementioned studies rely on augmentation/transformation of the original data, while in our proposed framework, explicit transformation is replaced by an internal operation.

\subsection{Adversarial Training} Goodfellow {\em  et al.} \cite{goodfellow2015explaining} were the first to propose adversarial regularisation as a training method \textcolor{black}{for} improving the robustness of the neural net model. However, computing adversarial samples is required to obtain the ground truth. Miyato {\em  et al.} \cite{DBLP:journals/corr/MiyatoMKNI15} proposed local distributional smoothness (LDS), an alternative approach to computing the adversarial perturbations without the requirement of target labels. Later, VAT \cite{8417973} was presented as a means for unsupervised average approximation of adversarial perturbations using multiple inputs. VAT is adopted in \textcolor{black}{the} clustering task \cite{Hu2017} to improve the model's smoothness and to impose the invariant representation. Instead of computing adversarial perturbation through the computation of gradient, Yang {\em  et al.} \cite{NEURIPS2020_6740526b} utilise an additional module for generating adversarial perturbation with the aim of improving the robustness \textcolor{black}{of} the clustering process. In our work, we introduce VAT as a straightforward technique of computing the gradient with the model's predictions only, rather than the target distributions. Furthermore, VAT \cite{8417973} demonstrates the ability to be applied to a variety of data types. For example, in \cite{9451540}, VAT is successfully applied to non-normalized features extracted by a GAN framework. However, our goal is to learn the representation and not to cluster the input data. Hence, this work computes multiple instances with the help of VAT and enforces invariant predictions \textcolor{black}{on} the method.

\section{Method}
\label{section:method}

\begin{figure}[t]
    \centering
    \includegraphics[width=0.5\textwidth]{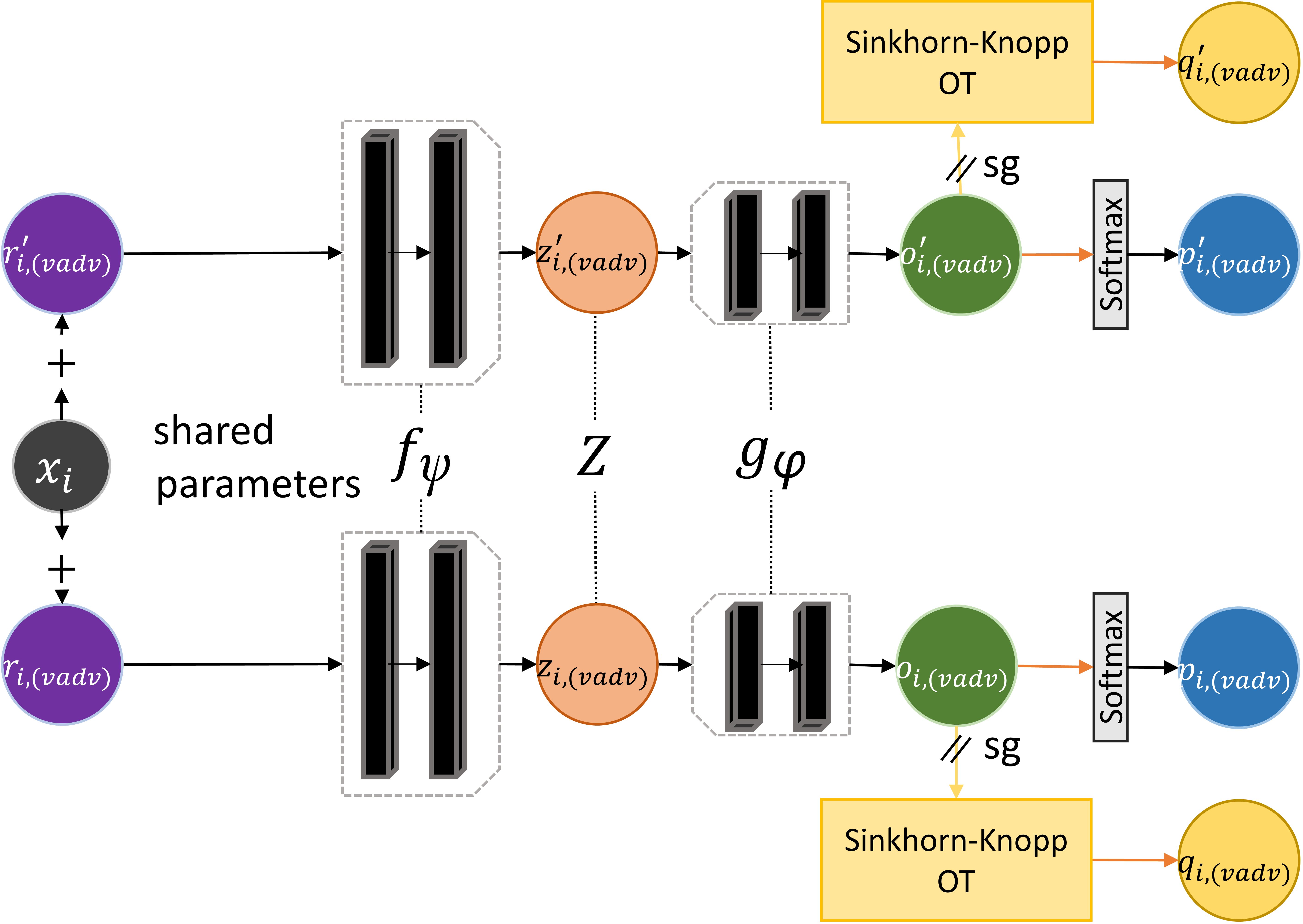}
    \caption{This diagram illustrates our proposed framework with the two implemented modules. The encoder part, denoted by ($f_\psi$), projects the input ($x_i$), augmented by VAT perturbations ($r_{i,(vadv)}$), onto the latent representation ($z_{i,(vadv)}$). The classifier module ($g_\phi$) outputs a probabilistic output ($p_{i,(vadv)}$). Lastly, based on the computed logit predictions ($o_{i,(vadv)}$), model predictions prior to the Softmax function), the Sinkhorn-Knopp \cite{NIPS2013_af21d0c9} algorithm generates the desired target distribution ($q_{i,(vadv)}$). Note that in the diagram, the prime superscript index (e.g., ${p'}_{i,(vadv)}$) denotes the second augmented instance by the VAT process from the same data point  ($x_i$)}.
    \label{fig:framework}
\end{figure}

In this section, we describe the training process of our method. Figure \ref{fig:framework} depicts the proposed framework, which consists of two trainable neural net modules: 1) an encoder module $(f_\psi)$ with parameters $\psi$ that projects the data to a bottleneck latent representation (also known as embedding space); and 2) a classifier module ($g_\phi$) with parameters $\phi$ that maps the latent representations to a discrete probabilistic distribution. \par

Let the data be an unlabelled collection of data points $\mathcal{X} = \{x_i\}_{i=1}^n$, where $x_i^T \in \mathbb{R}^{d}$ is a \textit{d}-dimensional vector. Each data point $x_i$ is associated with a distribution $p_i \in \Sigma_k$ that belongs to the probability simplex $\Sigma_k \myeq \{ \pi \in \mathbb{R}^{k}_{+} : \sum_{j=1}^k \pi_{j}=1\}$, and $k$ is a predefined hyper-parameter that indicates the length of the distribution. Our goal is to train a neural net framework in \textcolor{black}{an} unsupervised way, in which the encoder projects the data to the bottleneck representation $z_i = f_{\psi}(x_i) \in \mathbb{R}^{l}$, where $l$ is a hyper-parameter indicating the dimension of the embedding space. Then the classifier maps these representations to $o_i= g_{\phi}(z_i) \in \mathbb{R}^{k}$ (known as \textit{logits}) \textcolor{black}{at first, and then} to the associated distribution $p_i = \textit{Softmax} (o_i)$. For simplicity, hereafter, we denote the parameters of the complete framework as $\theta = \{\psi, \phi\}$ \textcolor{black}{, which represents the model $$p_i = E_\theta(x_i) = \textit{Softmax}(g_\phi(f_\psi(x_i))) ~.$$} \par

In supervised learning, the classification paradigm of a neural net is typically optimised by utilising stochastic gradient descent (SGD) to minimise an entropy loss function. In our implementation, we also use SGD to minimise the Kullback–Leibler divergence (KLD), also referred to as relative entropy, which measures the difference between two arbitrary discrete probabilistic distributions $\mathbf{\Pi} = \{\pi_i\}_{i = 1}^m \in (\Sigma_k)^m$\text{,} $\mathbf{\Gamma} = \{\gamma_i\}_{i = 1}^m \in (\Sigma_k)^m$ and is defined as,
\begin{equation}
\begin{split}
\mathcal{D}_{KL}(\mathbf{\Pi} \vert \vert \mathbf{\Gamma}) &= \frac{1}{m} \sum_{i=1}^m \sum_{j=1}^k \left[ \pi_{i, j} \text{log}\left(\frac{\pi_{i,j}}{\gamma_{i,j}}\right)\right] \\
&= \frac{1}{m} \sum_{i=1}^m \sum_{j=1}^k  \left[ \pi_{i,j} \text{log}(\pi_{i,j}) - \pi_{i,j} \text{log}(\gamma_{i,j}) \right] \\ 
&= \frac{1}{m} \left(-H(\mathbf{\Pi}) + H(\mathbf{\Pi},\mathbf{\Gamma})\right)
\text{ \ ,} 
\label{eq:entropy objective}
\end{split}
\end{equation}
where, $H(\mathbf{\Pi}, \mathbf{\Gamma})= -\sum_i \sum_j \pi_{i, j} \log(\gamma_{i, j})$ and $H({\mathbf{\Pi}}) = H(\mathbf{\Pi}, \mathbf{\Pi})$ is the entropy function, where $m \in \mathbb{N}$ and $ m > 0$.
\par

An objective function in Eq. (\ref{eq:entropy objective}) requires the target distribution. In unsupervised learning, these distributions are not available, and hence the problem is intractable. In the following, we firstly describe the process of generating transformed instances from the same data point with the aim of learning invariant representations. To achieve the desired invariance, two transformed instances are generated by repeating the self-transformation process described in Subsection \ref{section:vat}. The self-transformation can be seen as an internal mechanism based on VAT \cite{8417973}, which only requires the training samples as the input and relies on two hyper-parameters ($\epsilon$ and $\xi$). Secondly, with the help of the Sinkhorn-Knopp algorithm \cite{NIPS2013_af21d0c9}\textcolor{black}{,} in order to prevent degenerate solutions, we leverage the predictions made by the proposed framework in the transformed instances and compute two target distributions. Ultimately, the full training objective, presented in Subsection  \ref{section:detailed method}, is used to optimise the framework in a contrastive fashion, similar to supervised classification paradigms. The implementation of the Sinkhorn-Knopp algorithm \cite{NIPS2013_af21d0c9} requires to fix a hyper-parameter which controls the smoothness of the computed target distributions.
By assigning a constant value to the Sinkhorn-Knopp $\lambda$-parameter, the classification module will gradually increase the sparsity of the performed predictions, i.e., convert the framework's predictions into a categorical distribution, resulting in a sub-grouping effect in the embedding space. We therefore introduce an adaptation policy in Sub-section \ref{section:preventing segmentation} to adjust the smoothness of the target distribution by dynamically preserving the entropy value of the computed distributions to a defined level. Lastly, the overall structure of the training process is described in Algorithm \ref{alg:training}.

\subsection{Imposing Invariance in Representation}
\label{section:vat}

To learn discriminative representations of the training data without providing explicit target distributions (i.e., labels), we require a mechanism to impose invariant representations \textcolor{black}{on} arbitrary transformations. We achieve this invariance by selecting a flexible internal mechanism and avoiding processing functions for specific data structure\textcolor{black}{,} such as visual transformation. That is, we modify and implement a contrastive form of VAT \cite{8417973} since it does not require explicit definition of perturbations and therefore can be applied widely to a range of data types. Additionally, VAT \cite{8417973} perturbs the input data without requiring the underlying target distribution, making it suited for our proposed unsupervised method. \par

To begin with, we denote the adopted VAT self-transformation process as $T_{\theta^{(s)}}()$, derived from the framework parameters $\theta^{(s)}$ at training step $s$, with a training sample $x_i$ such as $x_{i,(vadv)}^{(s)} = T_{\theta^{(s)}}(x_i)$, which outputs the corresponding transformed instance $x_{i,(vadv)}^{(s)}$. This self-transformation mechanism, $T_{\theta^{(s)}}()$, relies only on two predefined hyper-parameters, $\epsilon$ and $\xi$. In the following, we describe the two steps of VAT, namely the forward and backward propagation\textcolor{black}{,} in estimating VAT perturbations:
\begin{equation}
\begin{split}
    r_{i,(vadv)}^{(s)} \approx	\epsilon \frac{\mathrm{g}_{i}^{(s)}}{\Vert \mathrm{g}_i^{(s)}\Vert^2} \text{ \  , \ \ \ \ \ } \\ \text{ \ \ where \  \ } \mathrm{g}_i^{(s)}= \nabla_{r^{(s)}_i} \mathcal{D}_{KL}\left(p_i^{(s)}\vert \vert p_{i,(r)}^{(s)}\right)
\label{eq:radv}  
\end{split}
\end{equation}
where $p_{i,(r)}^{(s)} = E_{\theta^{(s)}}(x_i + \xi \cdot r_i/\Vert r_i\Vert^2)$ denotes the forward propagation (first step) of a single point $x_i$, with $\mathbf{P}_r^{(s)} = \{p_{i,(r)}^{(s)}\}_{i = 1}^m \in (\Sigma_k)^m$ indicating the perturbed model predictions of a mini-batch at each step $s$, and the random perturbations of a mini-batch $R = \{r _{i}\}_{i=1}^m \in (\mathbb{R}^d)^m$, respectively, being the same length as the input dimension sampled from an i.i.d. normal distribution $N(0,1)$. Each vector $r_{i}$ is normalised by its Euclidean norm, and $\xi > 0 $ is a scaling factor that weights the impact of random perturbations. For the backward propagation (second step), $\mathrm{g}_i$ is the computed gradient with respect to the random perturbation $r_i$, and accordingly, $r_{i,(vadv)}^{(s)}$ is the normalised weighted gradient. The two hyper-parameter scalars $\xi$ and $\epsilon$ regularise the corresponding perturbations to the defined upper $\ell_2$-norm level. Note that Eq. (\ref{eq:radv}) refers to the VAT formulation for an individual data point of the $i$-th index. However, multiple elements can also be applied with the same formulation by averaging the mini-batch. For convenience, in the remainder of this section we denote the two steps of \textcolor{black}{the} VAT process as  $$T_{\theta^{(s)}}(x_i) = x_{i,(vadv)}^{(s)} = x_i + r^{(s)}_{i,(vadv)} $$ \par

\subsection{Self-Labelling Process}
\label{section:detailed method}

Given a mini-batch of data $\{x_i\}_{i = 1}^{m}$ of size $m$ and a current value for the model parameters \textcolor{black}{$\theta^{(s)}$ at training step $s$}, we describe the learning of target distributions resembling the model output while separating the mini-batch into classes of the same size.
$\mathbf{P}^{(s)} = \{p^{(s)}_i\}_{i = 1}^m \in (\Sigma_k)^m$ \textcolor{black}{denotes} the outputs for the model with the current parameter \textcolor{black}{$\theta^{(s)}$}.
The target distributions $\mathbf{Q}^{(s)} = \{q^{(s)}_i\}_{i = 1}^m \in (\Sigma_k)^m$ are derived by minimising a transport cost.
In turn, the loss function used to update parameter $\theta^{(s+1)}$ depends on $\mathbf{Q}^{(s)}$. \par

By using the model's predictions, $\mathbf{P}^{(s)}$ \textcolor{black}{of each training step ($s$)}, directly as an unconstrained target distribution \textcolor{black}{may} result in degenerate solutions in which all \textcolor{black}{training elements} are assigned to the same \textcolor{black}{output} index. Motivated by the studies of Asano {\em  et al.} \cite{asano2020self} and Caron {\em  et al.} \cite{caron2020unsupervised}, we compute the target distributions $\mathbf{Q}^{(s)}$ with the help of a self-labelling process.  We derive self-labels from the model's predictions \textcolor{black}{$\mathbf{P}^{(s)}$} under two different constraints in order to avoid degenerate solutions. Specifically, we avoid degenerate solutions by 1) regularising  the distribution of the target's marginal across all $k$ outputs ($\frac{1}{m} \sum_{i=1}^{m} q_{i,j}^{(s)} = \frac{1}{k}\mathbbm{1}_{k}$) and 2) through an additional entropy constraint to control the sparsity of the \textcolor{black}{computed} target distributions. The first constraint ensures an equal population \textcolor{black}{for} each index of the probabilistic distribution, preventing all data points from being assigned to a single index. The second condition, on the other hand, excludes solutions where all training samples are embedded at a constant value in the latent representation. The target $\mathbf{Q}^{(s)}$ can be efficiently solved as a regularised optimal transport plan via the application of the Sinkhorn-Knopp algorithm \cite{NIPS2013_af21d0c9}, which has a solution in the form
\begin{equation}
\begin{split}
\mathcal{U}(b,c) = \{\Pi \in (\Sigma_k)^{m}\vert \Pi \mathbbm{1}_m = b, \Pi^T\mathbbm{1}_k=c\} \text{ ,} \\ \text{where , } b= \frac{1}{m}\mathbbm{1}_{m} \text{ \ , \ } c=\frac{1}{k}\mathbbm{1}_{k} \text{ \ , \ \ \ \ \ }
\label{eq:q definition}
\end{split}
\end{equation}
where $b$ and $c$ denote the two uniform marginals of the first constraint. In fact, the required target matrix belongs to the transportation polytope 
\begin{equation}
\mathbf{d}^{\lambda}_{\mathbf{P}}(b,c) := \mymin\limits_{\Pi \in \mathcal{U}(b,c)}\left<\Pi,-\log \left(\mathbf{P}\right) \right> - \lambda H(\Pi) \text{ \ ,}
\label{eq:sinkhorn}
\end{equation}
where $\left< \cdot, \cdot \right>$ indicates the inner product of the two matrices, and the problem is formulated with a weighted entropy constraint, $\lambda H(\Pi)$, which also regularises the linear program leading to a unique solution \cite{peyreot}, $\lambda > 0$ is a weighting scalar. The role of the weighted entropy constraint is to control the smoothness of the solution $\mathbf{Q}^{(s)}$ (we will discuss the impact of $\lambda$ later in Section \ref{section:preventing segmentation}). $\mathbf{O}^{(s)} = \{o^{(s)}_i\}_{i = 1}^m \in \mathbb{R}^{k\times m}$ denotes the model \textit{logit} predictions of the corresponding mini-batch \textcolor{black}{at the training step $s$}, $\mathbfcal{K}^{(s)} = \textit{Softmax} \left( \frac{\mathbf{O}^{(s)}}{\lambda}\right)$.
Then, the target $\mathbf{Q}^{(s)}$ is computed as follows.
Set $\nu^{(0)}=\mathbbm{1}_k$, $\upsilon^{(0)} = b / (\mathbfcal{K}^{(s)} \nu^{(0)})$, and $T \in \mathbb{N}$ to be a fixed number of iterations.
Then for any $t \in \{0, \hdots T - 1\}$
\begin{equation}
\begin{split}
(\upsilon^{(t + 1)},\nu^{(t + 1)}) \leftarrow (b/({\mathbfcal{K}^{(s)}}\nu^{(t)}),c/({\mathbfcal{K}^{(s)}}^{\top}\upsilon^{(t)})) \text{, \ \ and,}\\
\mathbf{Q}^{(s)} = \mathbf{diag}(\upsilon^{(T)}) \, \mathbfcal{K}^{(s)} \, \mathbf{diag}(\nu^{(T)})
\label{eq:sinkhorn 2}
\end{split}
\end{equation}
$\upsilon$, $\nu$ are updated at each iteration ($t$) of the Sinkhorn-Knopp algorithm \cite{NIPS2013_af21d0c9} until a predefined tolerance or the maximum number of iterations is reached (\textcolor{black}{for convenience, the full implementation, including iterative steps} of Eq. \ref{eq:sinkhorn 2} is later presented in Algorithm \ref{alg:sinkhorn}). For a detailed explanation of the Sinkhorn-Knopp algorithm, we refer readers to \cite{NIPS2013_af21d0c9,peyreot}. \par

Neural network models demonstrate sensitivity to adversarial perturbations, which is used to increase their robustness \cite{goodfellow2015explaining}. Aiming to achieve invariant representation, we replace image transformation functions and leverage the internal transformation mechanism derived from the VAT process. To achieve this, we modify Eq. (\ref{eq:entropy objective}) based on a swapping form, as proposed in \cite{caron2020unsupervised}, and integrate it \textcolor{black}{into} our proposed model by computing multiple augmented instances via the VAT process for each data point. By computing and exchanging the corresponding target distributions of the transformed instances, the model is encouraged to make invariant prediction\textcolor{black}{s} in the presence of adversarial perturbations\textcolor{black}{,} with the final training objective being defined as
\begin{equation}
\begin{split}
L_{\theta}(X,\theta^{(s)}) &= \mathcal{D}_{KL}(\mathbf{Q'}_{(vadv)}^{(s)}\vert \vert \mathbf{P}_{(vadv)}^{(s)}) \\ &+ \mathcal{D}_{KL}(\mathbf{Q}_{(vadv)}^{(s)}\vert \vert \mathbf{P'}^{(s)}_{(vadv)}) \text{ ,}
\label{eq:final eqadv}
\end{split}
\end{equation}

\begin{figure}[t]
    \centering
    \includegraphics[width=0.5 \textwidth]{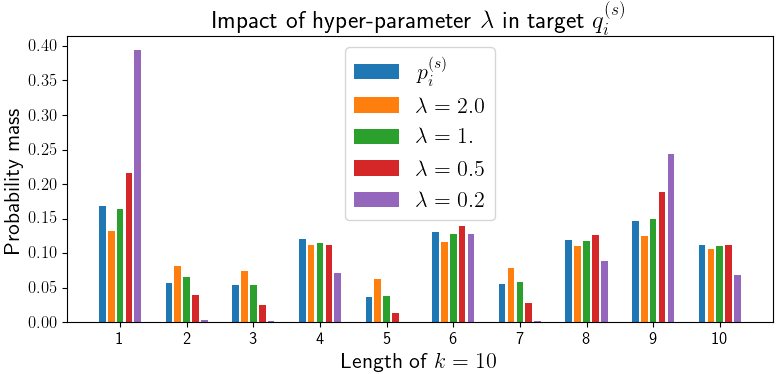}
    \caption{An illustration of the distribution of the probability mass of a single prediction $p_i^{(s)}$ at step $s$, and the corresponding distribution of the target $q_i^{(s)}$ (computed through the Sinkhorn-Knopp algorithm \cite{NIPS2013_af21d0c9}), by varying the value of parameter $\lambda$. Note that the target distributions are computed from the same mini-batch prediction $\mathbf{P}^{(s)}$.}
   \label{fig:plot p qs}
\end{figure}

\begin{figure}[t]
    \centering
    \includegraphics[width=0.5 \textwidth]{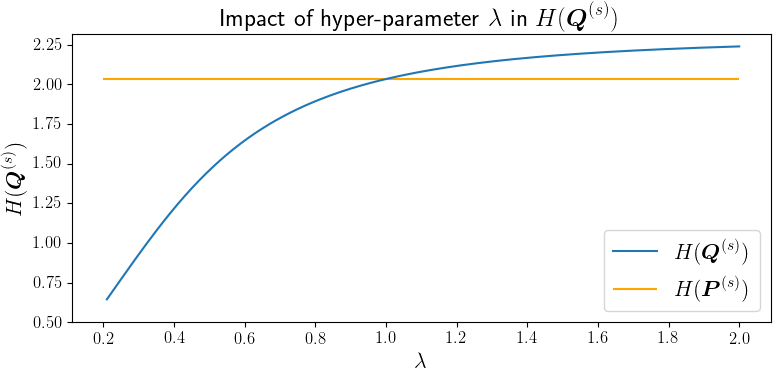}
    \caption{The plot shows the value of the regularised entropy term $H(\textbf{Q^{(s)}})$, at $s$ training step, by varying $\lambda$ over a wide interval. The target $\textbf{Q}^{(s)}$ is computed from the same predictions across the range of $\lambda$ (x-axis). For clarity, the orange horizontal line indicates the entropy value of the original $\mathbf{P}^{(s)}$.}
   \label{fig: hq loss}
\end{figure}
where \textcolor{black}{$\mathbf{P}_{(vadv)}^{(s)} = \{p_{i,(vadv)}^{(s)}\}_{i=0}^{m} \in $ and $p_{i,(vadv)}^{(s)} = E_{\theta^{(s)}}(T_{\theta^{(s)}}(x_i))$} denotes a model prediction obtained from an augmented instance via the self-transformation mechanism of VAT. For convenience, the prime superscript denotes the corresponding predictions \textcolor{black}{($\mathbf{P'}_{(vadv)}^{(s)} = E_{\theta^{(s)}}(T_{\theta^{(s)}}(x_i))$)} and the target distributions, $\mathbf{Q'}_{(vadv)}^{(s)}$, which are made by the proposed framework for the same data points by re-sampling the perturbations $r_i$ and repeating the same steps of forward and backward propagation ($T_{\theta^{(s)}}()$, Eq. \ref{eq:radv}), respectively.
Lastly, $\mathbf{Q'}_{(vadv)}^{(s)}$ and $\mathbf{Q}_{(vadv)}^{(s)}$ denote the target distribution derived from the corresponding perturbed model predictions $\mathbf{P'}^{(s)}_{(vadv)}$ \textcolor{black}{and} $\mathbf{P}^{(s)}_{(vadv)}$ at each training step $s$. We proceed with the computation of the targets, $\mathbf{Q'}_{(vadv)}^{(s)}$ and $\mathbf{Q}_{(vadv)}^{(s)}$, per mini-batch individually. All transformation instances and computations of the target distributions are computed individually at the corresponding training step, and the model's parameters are updated accordingly using SGD
\begin{equation}
\theta^{(s + 1)} 	\leftarrow \theta^{(s)} - 	\eta_\theta \nabla_\theta L_{\theta}(X,\theta^{(s)}),
\label{eq:updating theta}
\end{equation}
where $\eta_\theta$ is the positive learning rate.

In this work, the encoder is meant to identify common patterns in the data by optimising the framework parameters with the proposed objective function, allowing semantically similar data to reside in the same spatial neighbourhood in the latent representation ($z_i \in Z$).\par

\subsection{Adjusting the Smoothness of the Target Distribution}
\label{section:preventing segmentation}

\textcolor{black}{At} this stage, we briefly recall the formulation of the Sinkhorn-Knopp algorithm \cite{NIPS2013_af21d0c9} and the impact of the weighted entropy constraint, $\lambda H(\Pi)$, as expressed in Eq. (\ref{eq:sinkhorn 2}). The proposed adaptation policy assigns the corresponding value of hyper-parameter $\lambda$ dynamically based on the smoothness of the computed target distribution, $H(\mathbf{Q}^{(s)})$, \textcolor{black}{at the corresponding training step $s$} related to the warmup iterations ($T_{warm}$) and the defined target entropy ($h_{target}$). \par

\SetKwInOut{Initialize}{Initialize}
\SetKwInOut{Generate}{Generate}

\begin{algorithm}[t]
\SetKwFunction{FSinkhorn}{Sinkhorn}
    \SetKwProg{Fn}{Function}{:}{}
    \Fn{\FSinkhorn{$\mathbfcal{K}, \ T = 3$}}{
        $r= \frac{1}{m}1_{m}$ \text{, } $c=\frac{1}{k}1_{k}$ \text{, } $\nu=1_k$ \\
        \For{$t<T$}{
            $u = r / (\mathbfcal{K} \cdot v)$ \textcolor{gray}{\hspace{0.1mm} \% First step $u \in  \mathbb{R}^{m}$}\\  
            $v = c / (\mathbfcal{K}^T \cdot u)$ \textcolor{gray}{\hspace{0.1mm} \% Second step $v \in \mathbb{R}^{k}$}
        }
        $\mathbf{Q} = (\textit{diag}(v)^T \cdot (\mathbfcal{K} \cdot \textit{diag}(v)))$ \\
 \Return $ \mathbf{Q} \ / \ \sum_j^k \mathbf{Q}_{i,j}$ \textcolor{gray}{\hspace{0.1mm} \% Normalize across $k$ dimension} 
}
\textbf{End Function}
\caption{Sinkhorn-Knopp Algorithm \cite{NIPS2013_af21d0c9}}
\label{alg:sinkhorn}
\end{algorithm}

\begin{algorithm}[t]
\SetKwFunction{FAdapting}{Adapting}
    \SetKwProg{Fn}{Function}{:}{}
    \Fn{\FAdapting{$O, \ \lambda, \ s, \ T_{\text{warm}} = 100 \times \frac{n}{m}$}}{ 
         $h_{\text{target}}=\text{log}(\sqrt k)$ \text{, } $h_{\text{tol}}=5*10^{-3}$ \text{, } $h_{\text{step}}=10^{-1}$ \\
         $h_{\text{scale}} = (cos(\pi * \textit{Min}( s / T_{\text{warm}} , \ 1.)) + 1) / 2$ \\
         $\widehat{h_{\text{target}}} = h_{\text{target}} + (\text{log}(k) - h_{\text{target}}) * h_{\text{scale}}$ \\
         $\mathbf{Q} = $ \FSinkhorn{$\textit{Softmax}(O/\lambda)$} \textcolor{gray}{\hspace{0.1mm} \% Compute $\mathbf{Q}$}\\
        \For{$t< 5$}{
            \uIf{$(H(\mathbf{Q})- h_{\text{target}}) > h_{\text{tol}}$}{
                $\lambda = \text{Max}(\lambda - h_{\text{step}}, 0.1)$ \textcolor{gray}{\hspace{0.1mm} \% Update  $\lambda$}
            }\uElseIf{($H(\mathbf{Q})- h_{\text{target}}) < -h_{\text{tol}}$}{
                $\lambda = \text{Min}(\lambda + h_{\text{step}}, \ 1)$ \textcolor{gray}{\hspace{0.1mm} \% Update  $\lambda$ }
            }\Else{
                Exit loop
            }
            $\mathbf{Q} = $ \FSinkhorn{$\textit{Softmax}(O/\lambda)$} \textcolor{gray}{\hspace{0.1mm} \% Update $\mathbf{Q}$}
        }
 \Return $ \mathbf{Q} \text{ \ , \ } \lambda$ \textcolor{gray}{\hspace{0.1mm} \% Return updated target and $\lambda$}
}
\textbf{End Function}
\caption{Adapting $\lambda$ Policy}
\label{alg:adapting}
\end{algorithm}

\begin{algorithm}[t]
\SetKwFunction{FVAT}{VAT}
    \SetKwProg{Fn}{Function}{:}{}
    \Fn{\FVAT{$\theta, B, \mathbf{P}$}}{
        $R \sim N(0,1)$ \textcolor{gray}{\hspace{0.1mm} \% Sample random perturbations from i.i.d Normal Distribution $R=\{r_i\}_{i=1}^m \in  (\mathbb{R}^{d})^m$}  \\ 
        \ForEach {$r_i \in R $}{
        $r_i \approx \xi * ( r_i / \Vert r_i\Vert^2 )$ \textcolor{gray}{\hspace{0.1mm} \% Normalize each row ($d$ dimension) of the random perturbations $\Vert r_i\Vert^2 \leq \xi$ }
        }
        $\mathbf{P}_r = \textit{Softmax}(g_\phi ( f_\psi(B + R))) $ \textcolor{gray}{\hspace{0.1mm} \% Model predictions}  \\ 
        $G= \nabla_{r} \mathcal{D}_{KL}\left(\mathbf{P}\vert \vert \mathbf{P}_r\right) $ \textcolor{gray}{\hspace{0.1mm} \% where $G=\{g_i\}_{i=1}^m \in  (\mathbb{R}^{d})^m$ the computed gradient of a mini-batch} \\ 
        $R_{(vadv)} = \{\}$ \textcolor{gray}{\hspace{0.1mm} \% define an empty set} \\
        \ForEach {$\mathrm{g}_i \in G $}{
        $r_{i,(vadv)} \approx \epsilon * ( \mathrm{g}_i / \Vert \mathrm{g}_i\Vert^2 )$ \textcolor{gray}{\hspace{0.1mm} \% Normalize each row ($d$ dimension) of the computed gradient $\Vert g_i\Vert^2 \leq \epsilon$ } \\
        $R_{(vadv)} := R_{(vadv)} \cup \{r_{i,(vadv)}\}$ 
        }
        $\mathbf{O}_{(vadv)} = g_\phi ( f_\psi(B + R_{(vadv)})) $ \textcolor{gray}{\hspace{0.1mm} \% Model predictions in VAT perturbed inputs prior to Softmax function (also called "logits''). Note $r_{i,(vadv)} \in R^{(vadv)}$.}  \\ 
        $\mathbf{P}_{(vadv)} = \textit{Softmax}(O_{(vadv)}) $ \textcolor{gray}{\hspace{0.1mm} \% Compute $\mathbf{P}$}  \\ 
 \Return $\mathbf{P}_{(vadv)} $, $O_{(vadv)}$
}
\textbf{End Function}
\caption{Computing VAT predictions}
\label{alg:vat}
\end{algorithm}

\begin{algorithm}[t]
\KwData{$X=\{x_{i}\}_{i=1}^{n}$}
\Initialize{$\theta^{(0)} = (\psi^{(0)}, \phi^{(0)}) \ , \lambda = 1, \ S = N$}
\For{$s \in \{0, \hdots S \}$}{
  \hspace{0.3mm}{$B^{(s)}=\{x_{1},...,x_{m}\}$}  \textcolor{gray}{\hspace{0.1mm} \% Sample mini-batch $B^{(s)} \in  \mathbb{R}^{m \times d}$}  \\
  \hspace{0.3mm}{ $\mathbf{P}^{(s)} = \textit{Softmax} (g_{\phi^{(s)}}(f_{\psi^{(s)}}(B^{(s)})))$ } \textcolor{gray}{ \% Model predictions $\mathbf{P}^{(s)}$ at current step}  \\ 
  \hspace{0.3mm}{$\mathbf{P}^{(s)}_{(vadv)}, \mathbf{O}^{(s)}_{(vadv)}$ = \FVAT{$B^{(s)}, \mathbf{P}^{(s)}$}} \textcolor{gray}{ \% First group of VAT perturbed instances}  \\ 
  \hspace{0.3mm}{$\mathbf{Q}^{(s)}_{(vadv)} \text{ \ , \ } \lambda = $ \FAdapting{$\mathbf{O}_{(vadv)}^{(s)}, \lambda, s$}} \textcolor{gray}{ \% Adapt $\lambda$, compute target $\bf{Q}^{(s)}_{(vadv)}$} \\
  \hspace{0.3mm}{${\mathbf{P}'}^{(s)}_{(vadv)}, {\mathbf{O}'}^{(s)}_{(vadv)}$ = \FVAT{$B^{(s)}, \mathbf{P}^{(s)}$}} \textcolor{gray}{ \% Second group of VAT perturbed instances}  \\ 
  \hspace{0.3mm}{${\mathbf{Q}'}^{(s)}_{(vadv)} \text{ \ , \ } \lambda = $ \FAdapting{${\mathbf{O}'}_{(vadv)}^{(s)}, \lambda, s$}} \textcolor{gray}{ \% Adapt $\lambda$, compute target ${\mathbf{Q}'}^{(s)}_{(vadv)}$} \\
\hspace{1.mm}{ \Comment{Update framework's parameters eq. (\ref{eq:final eqadv})} } 
\begin{align*}
\begin{split}
L_{\theta}(B^{(s)},\theta^{(s)}) &= \mathcal{D}_{KL}(\mathbf{Q'}_{(vadv)}^{(s)}\vert \vert \mathbf{P}^{(s)}_{(vadv)}) \\ &+ \mathcal{D}_{KL}(\mathbf{Q}_{(vadv)}^{(s)}\vert \vert \mathbf{P'}^{(s)}_{(vadv)}) \\ 
\end{split}
\end{align*}
\begin{align*}
\begin{split}
\theta^{(s + 1)} 	\leftarrow \theta^{(s)} - 	\eta_\theta \nabla_\theta L_{\theta}(B^{(s)},\theta^{(s)})\\ 
\end{split}
\end{align*}

}
\caption{Training Process of the Proposed Framework.}
\label{alg:training}
\end{algorithm}
By assigning a constant value in the range $\lambda \in [1,\infty)$, the target $\mathbf{Q}^{(s)}$ is restricted to a more homogeneous solution than the model predictions $\mathbf{P}^{(s)}$, conditional on the two marginal constraints ($b$ and $c$). Consequently, the target $\mathbf{Q}^{(s)}$ will discourage the model from maximising the similarities between augmented pairs, leading to trivial solutions. In \textcolor{black}{the} case $\lambda \in (0 , 1 )$, the impact of the regularisation term is decreased when generating a solution $\mathbf{Q}^{(s)}$ with a lower entropy value. This then leads to assigning higher probabilities to the corresponding index of the target $\bf{Q}^{(s)}$, thus gradually increasing the similarities between the augmented data points with its VAT perturbations.

To empirically verify the above hypothesis, we vary the value of $\lambda$ and observe the effect on the smoothness of the target $\textbf{Q}^{(s)}$, which is computed from predictions made by our model in a mini-batch $\mathbf{P}^{(s)}$. Figure \ref{fig:plot p qs} illustrates a single prediction, selected from the computed mini-batch, \textcolor{black}{and the corresponding target distributions}, with the associated $\lambda$ values. This validates that an assigned value \textcolor{black}{of} $\lambda > 1$ decreases the sparsity of the target distributions (flattening), while a $\lambda < 1$ increases the confidence of the target distributions. Figure \ref{fig: hq loss} plots the impact of parameter $\lambda \in [0.2,2]$ as a function of $H(\bm{Q}^{(s)})$, where the orange horizontal line shows the entropy value of the initial model predictions $H(\mathbf{P}^{(s)})$. \par

After a large number of training epochs with a constant $\lambda < 1$, the target $\bf{Q}^{(s)}$ will gradually become a categorical distribution (one-hot vectors) and the model will converge to group $n/k$ data points per output index. In order to avoid any unwanted grouping effects and encourage the model to only learn similar predictions between  perturbed instances of the same data point, we propose to adapt parameter $\lambda$. This adaptation is based on the difference between a target entropy value and the $H(\mathbf{Q}^{(s)})$ as
\begin{equation}
\begin{split}
\lambda := 
  \begin{cases}
    \text{Max}(\lambda - h_{\text{step}}, 0.1), & \text{if \ } (H(\mathbf{Q}^{(s)}) -\widehat{h_{\text{target}}}) > h_{\text{tol}} \\
    \text{Min}(\lambda + h_{\text{step}}, 1) , & \text{if \ } (H(\mathbf{Q}^{(s)}) - \widehat{h_{\text{target}}}) < -h_{\text{tol}} \\
    \lambda, & \text{otherwise}
  \end{cases}
\label{eq:adjacting lambda eqadv}
\end{split}
\end{equation}
where $h_{\text{tol}}$ denotes a tolerance hyper-parameter, $h_{\text{step}}$ is a small step size for varying $\lambda$ according to the difference with $\widehat{h_{\text{target}}}$. The target entropy value ($\widehat{h_{\text{target}}}$) is adapted via a cosine warm-up schedule
\begin{equation}
\begin{split}
\widehat{h_{\text{target}}} = h_{\text{target}} + (\text{log}(k) - h_{\text{target}}) * h_{\text{scale}} \text{ , \ \ \ \ \ \ \ \ \ } \\
\text{ where } h_{\text{scale}} = (\text{cos}(\pi * \text{Min}( s / T_{\text{warm}} , \ 1.)) + 1) / 2 \text{ \ \ \ , }
\end{split}
\end{equation}
where $h_{\text{target}}$ is a pre-defined parameter, $s$ represents the current training step, and $T_{\text{warm}}$ denotes the total number of warm-up period. Recall that $k$ is the length of the probabilistic output of the framework. We utilise a cosine warm-up plan rather than a linear increasing target value to avoid a high $\lambda$ value in early training epochs, as this may lead to premature highly confident assignments (see Fig.~\ref{fig:plot p qs}). We set all hyper-parameters to constant values of 1) $h_{\text{target}} = \text{log}(\sqrt k)$, with target being the half value of the initial entropy level $\text{log}(k)$; 2) $h_{\text{tol}} = 5\times10^{-3}$, as a reasonable difference between the two entropy values; 3) $h_{\text{step}} = 10^{-1}$, to be sufficient by observing the impact of factor $\lambda$ as a function of $H(\mathbf{Q})$ in Fig. \ref{fig: hq loss};  and 4) $T_{\text{warm}} = 10^2  \times \frac{n}{m}$ in order to avoid computing very sparse target distribution\textcolor{black}{s}, $\mathbf{Q}^{(s)}$, in very early training epoch (where $\frac{n}{m}$ the training steps per epoch). We show that by keeping these values constant, they work well across all experiments and that the adaptation policy successfully prevents sub-grouping effects. We iterate the adaptation policy for $\lambda$ in Eq. (\ref{eq:adjacting lambda eqadv}) for a maximum of five times for each computed target distribution. (Recall that with an assigned value of $h_{\text{step}} = 10^{-1}$, the maximum iteration number for the adaptation policy (Eq. (\ref{eq:adjacting lambda eqadv})) is $10$, thus we select the median number of five iterations). \par

More details of the Sinkhorn-Knopp algorithm, the policy for adaptation of $\lambda$, the computation of VAT perturbations, and the computation of the target  as well as the entire training process are presented in Algorithms (\ref{alg:sinkhorn}), (\ref{alg:adapting}), (\ref{alg:vat}), and  (\ref{alg:training}), respectively. Further discussion of parameter selection and the model's behaviour, including the efficacy of our proposed framework, is presented in Section \ref{section:experimental section}. Additionally, we demonstrate the role of each component and the influence of different values of the predefined parameters via a detailed ablation study and sensitivity analysis. 

\section{Experimental Section}
\label{section:experimental section}

In this subsection, we present our empirical studies to demonstrate the capabilities of our proposed embedding strategy and to understand the influence of each hyper-parameter on the performance of the proposed framework.  To do so, we evaluate our method against a broad range of benchmarks, including visual, text content, audio, and mass spectrometry imaging (MSI) datasets:

\begin{itemize}
  \item \textbf{E-MNIST} is a visual dataset consisting of gray-scale handwritten \textcolor{black}{characters and digits. We use the subset of digits} (10 classes) with images' resolution of $28\times28$. This set includes $240,000$ training and $40,000$ testing instances.
  
  \item \textbf{Fashion-MNIST} is a visual dataset including an equal number of \textcolor{black}{$60,000$ training and $10,000$} testing elements. However, it is considered a more complicated dataset associated with $10$ different clothing categories. Similar to E-MNIST, it holds the same image dimension.
  
  \item \textbf{Reuters} is text dataset containing  a collection of news stories in the English language. We evaluate all the benchmark methods in the comparison list for the four categories: corporate/industrial, government/social, markets, and economics. We randomly select $10,000$ stories from these categories subset, and the same data-points are used across all models. Note that the four aforementioned classes are unevenly populated, with the larger class containing $\approx 43\%$ of the training instances.
  
  \item \textbf{20news} is a text corpus containing $18,040$ English news posts on $20$ different topics. The topics' population is approximately balanced with the corresponding categories in this dataset.
  
  \item \textbf{FSDD} is an audio dataset of recordings of spoken digits in English pronunciations. The set consists of $3,000$ recordings from a total of six different speakers. All recordings are trimmed to reduce additional sounds and save at an 8kHz frequency.
  
  \item \textbf{Mass Spectrometry Imaging (MSI)} is an important approach to investigating the spatial distribution of molecular species. The data structure is composed of a 3-d datacube $(x,y,z)$, where the first two dimensions refer to the spatial location of the object and the third dimension (spectral) to the mass-to-charge ratio  (m/z). Each spectrum (m/z) typically contains high-dimensional vector (for example, $1024$). In this study, we experiment with a rat brain sample, a dataset previously discussed in \cite{msi}, consisting of $13,320$ points of interest (sample), while the remaining points ($7,305$) surrounding the object are manually removed (background). The MSI dataset is evaluated via a linear regression with target values representing a colour-full image obtained through a t-SNE visualisation in \cite{msi}. Note that each colour of the evaluation image corresponds to different regions of the rat brain.
  
\end{itemize}

Motivated by proposing a generic feature learning strategy applicable to various data types, \textcolor{black}{we evaluate our proposed algorithm in two neural network architectures i.e., a CNN and an MLP.} By doing this, we make sure that the proposed framework is not subject to limitations where these neural architectures cannot be used. More specifically, \textcolor{black}{we apply a CNN architecture for} the visual datasets of E-MNIST and Fashion-MNIST. The corresponding data points of text, audio, and MSI datasets are used as vectors with an MLP architecture, respectively. For all text datasets, the pre-processing follows what is presented in deep embedding clustering (DEC) \cite{pmlr-v48-xieb16} by counting all the words in the dataset and computing the \textit{tf-idf} for each text, respectively. We only select $2,000$ most popular words (excluding the stop-words) for each dataset to return the same dimension of each data-point ($2000\text{-d}$). Note that, due to the \textit{tf-idf} word indexing of the text content, all generated data points preserve a high sparsity distribution of less than $\approx 2\%$ of non-zero values on average. We preprocess the audio dataset by converting each audio signal into a fixed $2\text{-d}$ spectogram obtained via Fourier frequencies with a single instance dimension of $48\times48$, which we then reshape to fit a vector with a size of $2304\text{-d}$. The MSI dataset is binned between the (m/z) range of $[50,1000]$ to $1024$ bins. The spatial locations of the dataset (x,y) are transformed into a single dimension, with each point consisting of the corresponding spectra. All spectra are normalised to the $[0,1]$ interval based on the maximum value of the specific bin index among all spectra.
\textcolor{black}{All benchmarks are evaluated to the same data points with no} additional augmentation/transformation functions to any specific experiment or extra pre-processing methods.
\par

\subsection{Baseline Methods}
We demonstrate the enhanced encoding capacity of the proposed feature learning strategy and its learning properties by comparing it with \textcolor{black}{four} traditional dimension reduction methods\textcolor{black}{, one self-supervised strategy that is applicable to a broad range of data types,} and five variants of deep AEs: 1) principal components analysis (PCA) \cite{10.1162/089976699300016728}; 2) non-negative matrix factorization (NMF) \cite{2009nmf}; 3) Isomap embedding \cite{doi:10.1126/science.290.5500.2319}\textcolor{black}{, 4) local linear embedding (LLE) \cite{Roweis2000}; 5) contrastive predictive coding \cite{2018arXiv180703748V}; 6)} the vanilla AE \cite{HintonSalakhutdinov2006b}; 7) variational autoencoder (\textcolor{black}{VAE}) \cite{Kingma2014}; 8) adversarial autoencoder (AAE) \cite{makhzani2016adversarial}; 9) sparse autoencoder (SAE) \cite{sparse}; and 10) denoise autoencoder (DAE) \cite{10.1145/1390156.1390294}. For a fair comparison, we implement the same model architecture for all deep models based on the MLP \textcolor{black}{or CNN design except the CPC that a CNN architecture is used across all experiments due to its learning process}. Additionally, all benchmark methods are applicable to various data types and not specified to any particular data.

\subsection{Evaluation Metrics} Fair evaluations of the generated features are not a straightforward task, since different metrics may account for different perspectives. Here, we employ \textcolor{black}{five} different evaluation metrics to examine the effectiveness of the algorithms under comparison.

\begin{itemize}
  \item \textbf{Linear Evaluation}: This metric was first used in \cite{10.5555/3305381.3305435} and later adopted in a range of self-supervised learning tasks as \textcolor{black}{a} linear evaluation of the generated representations. In more detail, we freeze the encoder's parameters and train a single layer as a classifier on top of the embedding space of each model. This metric measures the linear separability of the representations via the classification performance \textcolor{black}{(accuracy)}, i.e., the difference between the classifier predictions and the ground truth.
  
  \item \textbf{Graph Evaluation}: Using the t-SNE \cite{vanDerMaaten2008} technique, we plot the extracted features in 2-dimensional space. Despite the fact that there is no simple method for validating the generated plots, we attempt to assess the spatial position of each point using the labels of the $21$-nearest-neighbor points. To accomplish this, we use a $k$-NN implementation and generate a histogram for each data point based on the ground truth by counting the labels of the closest points. \textcolor{black}{A positive prediction is made if and only if the label of the selected data point matches the highest index of its histogram.}  For a quantitative metric, we compute the proportion of positive predictions divided by the total number of points. 

  \item \textcolor{black}{\textbf{Mutual Information (MI)}}: \textcolor{black}{This evaluation metric measures the mutual dependency between the extracted features ($Z$), obtained from the bottleneck layer, and the given labels ($Y$). To achieve this, we implement a neural net architecture and apply the estimation loss function based on the proposed study of Mutual Information Neural Estimation (MINE) \cite{pmlr-v80-belghazi18a}. MINE provides the framework for measuring the MI for continuing variables. For better interpretation of the estimation, all labels are encoded as one-hot vectors. The larger metric value indicates the higher mutual dependency between the variables $I(Z;Y)$.}
    
  \item \textbf{Clustering}: A \textit{k}-means implementation is employed to cluster the generated features. To improve precision, we initialised the cluster centroids for $20$ independent rounds and only preserved the centroids with the lowest distances. Due to the stochastic initialization of the centroids, we select the highest measurement of the last $10$ training epochs for each method. We report three unsupervised metrics of this evaluation strategy: the clustering accuracy (ACC) based on the best mapping between the ground truth and the predictions of the \textit{k}-means algorithms, the normalised mutual information (NMI) and the adjusted rand index (ARI). Using \textit{k}-means as an evaluation method, we aim to investigate the separability of the cluster centres via an unsupervised approach.
  
  \item \textbf{Linear Regression}: This strategy is exclusively for the evaluation of the MSI dataset, which is an unlabeled dataset. Specifically, the corresponding model is used to project each spectrum in the dataset into the ($z$) dimension. Afterwards, we freeze the model parameters and a single linear layer is trained with the extracted features as the input and the three RGB values obtained through the t-SNE embedding strategy reported in \cite{msi} as the target output, where each individual colour refers to a specific region. The final quantity measurement is derived by summing all squared differences (residual sum of squares (RSS)) between the target values and the predictions of linear regression.

\end{itemize}

\subsection{Implementation Details} 

In this sub-section we present the implementation details of the compared algorithms. These details are used in most of the experiments unless otherwise stated. All compared methods contain the same number of hidden layers and units, as well as the same dimension in embedding space in each experiment. More specifically, \textcolor{black}{the CNN architecture for the encoder used for the vision dataset consists of three layers. The first layer uses 128 filters. Every other layer contains double the number of filters. The kernel size is set to five and the stride to two. The same architecture is used across all AE frameworks, and the decoder part holds the same inverse structure. In the CPC model, the stride is set to one since multiple smaller patches are generated from the same image.} \par
\textcolor{black}{On the other hand, across all text, audio and MSI training datasets, we employ an MLP architecture encoder consisting of two hidden layers, and each layer contains  $1024$ units. The architecture of the classifier module ($g_\phi$) in both the MLP and CNN} consists of two hidden layers, each layer containing $128$ units, and the output layer has $k=100$ units. The effects of different dimensions in the probabilistic output ($k$) are also presented in the study. \textcolor{black}{Across all experiments and models, we set} the embedding dimension to $l=10$ units. For our proposed method, the batch normalisation layers (BN) \cite{pmlr-v37-ioffe15} are included in all experiments. We find, however, that the vanilla AE achieves better results without including BN layers. We therefore conduct experiments with AEs with or without BN across all methods. The best captured performance is reported for each method. The decoder part of the autoencoder frameworks has the same architecture as the encoder part. \par

\textcolor{black}{In Table \ref{tab: parameters}, we list the number of hyper-parameters and the training time for a full training round for the F-MNIST (CNN architecture) and MSI (MLP architecture) datasets, respectively, for each framework. Note that in CPC, we used a CNN implementation across all experiments due to the training process of using patches or temporal sequence of the input data point. However, for a fair comparison, we preserve the same or a larger number of parameters.}

\begin{table}[!h]
\centering

\caption{\textcolor{black}{The number of parameters for each framework under comparison and the training time (in seconds) for a full round.}}

\begin{tabularx}{250.pt}{l|c|c|c|c}

\cline{1-5}
\multirow{2}{*}{}& \multicolumn{2}{c}{\textbf{CNN}} & \multicolumn{2}{c}{\textbf{MLP}}  \\
\cline{2-5}
& \textbf{No. params}&\textbf{Training time}&\textbf{No. params}&\textbf{Training time}\\ \cline{1-5}

AEs&1.23M&8.1 sec/epoch&4.21M&0.5 sec/epoch\\
CPC&1.27M&12.7 sec/epoch&3.85M&7.9 sec/epoch\\
Ours&0.90M&12.9 sec/epoch&2.11M&1.6 sec/epoch\\
\cline{1-5}
\end{tabularx}
\label{tab: parameters}
\end{table}
\textcolor{black}{Note that all experiments are conducted in Python (v3.7.1) scripts. All neural net architectures are implemented under the PyTorch (v1.60) framework, and all reduction methods are implemented via the Scikit-Learn (v0.23.2) package. The training of all neural network models is accomplished on a GPU 'GeForce RTX 2080 Ti' with 12GB of memory and Cuda version 10.1. Additionally, the machine, to execute each of the training, was allocated with RAM of 12GB respectively.} \par

\begin{table*}[t]
\caption{This table includes the experimental results of two evaluation strategies. 1) \textcolor{black}{The accuracy of the} linear evaluation, and 2) \textcolor{black}{The accuracy measured based on the 21-nearest neighbor of the} graph evaluation. The top three results are highlighted. These results are the average of five independent runs with their standard deviation, respectively. Note that the labels are only used to classify and validate the extracted features.}
\begin{center}
\setlength\tabcolsep{4.pt}
\begin{tabular}{l||c|c|c|c|c||c|c|c|c|c}
\hline

\multirow{2}{*}{\shortstack[l]{\textbf{Method/} \\ \textbf{Dataset}}}& \multicolumn{5}{c||}{\textbf{Linear Classifier \textcolor{black}{(Accuracy)}}}  & \multicolumn{5}{c}{\textbf{Graph Evaluation (\textcolor{black}{The accuracy of the} 21-NN points) }} \\
\cline{2-11}
& \textbf{E-MNIST}&\textbf{F-MNIST}&\textbf{Reuters}&\textbf{20news}&\textbf{FSDD}& \textbf{E-MNIST}&\textbf{F-MNIST}&\textbf{Reuters}&\textbf{20news}&\textbf{FSDD}\\ \hline

PCA&82.9 $\pm$(0.1)&74.7 $\pm$(0.1)&
\setlength{\fboxsep}{0pt}\colorbox{silver}{88.8 $\pm$(0.1)}&46.5  $\pm$(0.3)&75.9  $\pm$(0.1)& 93.3 $\pm$(0.1)
&	77.3  $\pm$(0.1)&	92.5  $\pm$(0.1)&	47.0  $\pm$(0.3)&	79.9  $\pm$(0.6)
\\

NMF&81.1 $\pm$(0.1)&73.3  $\pm$(0.1)&85.6  $\pm$(0.1)&41.5  $\pm$(0.1)&76.9  $\pm$(0.1)&87.7 $\pm$(0.1)
&	76.0  $\pm$(0.1)&	90.3  $\pm$(0.1)&	43.4  $\pm$(0.2)&	86.2  $\pm$(0.3)
\\
LLE&N/A&53.9 $\pm$(3.7)&85.3 $\pm$(0.9)&\setlength{\fboxsep}{0pt}\colorbox{yellow}{61.5  $\pm$(1.6)}&64.4 $\pm$(0.2)&N/A&54.0 $\pm$(3.3)&90.4 $\pm$(0.1)&58.9 $\pm$(1.2)&81.8 $\pm$(0.4)\\

ISOMAP&N/A&76.3 $\pm$(0.1)&87.4 $\pm$(0.2)&50.8 $\pm$(0.4)&72.9 $\pm$(0.2)&N/A &79.2 $\pm$(0.2)&90.6 $\pm$(0.1)&54.2 $\pm$(0.2)&83.2 $\pm$(0.4)\\

AE&\setlength{\fboxsep}{0pt}\colorbox{brown}{97.3 $\pm$(0.1)}&
\setlength{\fboxsep}{0pt}\colorbox{brown}{78.3 $\pm$(0.6)}&
87.2  $\pm$(0.9)&
\setlength{\fboxsep}{0pt}\colorbox{brown}{58.3 $\pm$(0.3)}&
78.5 $\pm$(1.3)&
\setlength{\fboxsep}{0pt}\colorbox{brown}{97.3 $\pm$(0.1)}&
\setlength{\fboxsep}{0pt}\colorbox{silver}{81.6 $\pm$(0.3)}
&
\setlength{\fboxsep}{0pt}\colorbox{brown}{92.8 $\pm$(0.3)}&
\setlength{\fboxsep}{0pt}\colorbox{silver}{61.1 $\pm$(0.2)}&
\setlength{\fboxsep}{0pt}\colorbox{silver}{91.9  $\pm$(0.2)}
\\

AAE&95.8 $\pm$(0.1)&76.2 $\pm$(0.6)&87.2 $\pm$(0.8)&54.7 $\pm$(1.2)&69.6 $\pm$(2.3)&94.4 $\pm$(0.1)& 
\setlength{\fboxsep}{0pt}\colorbox{white}{80.2 $\pm$(0.4)}
&
\setlength{\fboxsep}{0pt}\colorbox{silver}{93.1 $\pm$(0.3)}&
\setlength{\fboxsep}{0pt}\colorbox{silver}{61.1 $\pm$(0.3)}&89.1 $\pm$(1.9)
\\

SAE&\setlength{\fboxsep}{0pt}\colorbox{brown}{97.3 $\pm$(0.1)}&78.2  $\pm$(0.2)&
\setlength{\fboxsep}{0pt}\colorbox{brown}{88.0 $\pm$(0.9)}&
56.6 $\pm$(0.2)&77.6  $\pm$(1.1)&\setlength{\fboxsep}{0pt}\colorbox{brown}{97.3  $\pm$(0.1)}&
\setlength{\fboxsep}{0pt}\colorbox{brown}{80.8 $\pm$(0.4)}
&	91.8  $\pm$(0.3)&	57.6  $\pm$(0.4)&	87.8  $\pm$(0.9)
\\

DAE&\setlength{\fboxsep}{0pt}\colorbox{silver}{97.4 $\pm$(0.1)}&
\setlength{\fboxsep}{0pt}\colorbox{silver}{79.0 $\pm$(0.5)}&
87.3 $\pm$(0.8)&
\setlength{\fboxsep}{0pt}\colorbox{white}{58.2 $\pm$(0.4)}&
\setlength{\fboxsep}{0pt}\colorbox{silver}{78.8 $\pm$(1.3)}&
\setlength{\fboxsep}{0pt}\colorbox{silver}{97.4 $\pm$(0.1)}&
\setlength{\fboxsep}{0pt}\colorbox{yellow}{81.9 $\pm$(0.4)}
&
\setlength{\fboxsep}{0pt}\colorbox{brown}{92.8 $\pm$(0.4)}&
\setlength{\fboxsep}{0pt}\colorbox{brown}{60.7 $\pm$(0.4)}&
\setlength{\fboxsep}{0pt}\colorbox{yellow}{92.3 $\pm$(0.5)}
\\

VAE&94.2  $\pm$(0.5)&72.2  $\pm$(1.6)&73.8  $\pm$(0.7)&
\setlength{\fboxsep}{0pt}\colorbox{white}{58.2 $\pm$(0.6)}&63.9  $\pm$(1.3)&95.0 $\pm$(0.3)&
\setlength{\fboxsep}{0pt}\colorbox{white}{78.8 $\pm$(0.5)}
&
\setlength{\fboxsep}{0pt}\colorbox{brown}{92.8 $\pm$(0.1)}&
\setlength{\fboxsep}{0pt}\colorbox{yellow}{61.6 $\pm$(0.2)}& 89.2 $\pm$(1.4)\\

CPC &82.8  $\pm$(1.0)&77.4  $\pm$(0.7)&85.5 $\pm$(0.5)&47.7 $\pm$(0.8)&\setlength{\fboxsep}{0pt}\colorbox{brown}{78.6  $\pm$(1.1)}&68.2 $\pm$(1.1)&
\setlength{\fboxsep}{0pt}\colorbox{white}{75.8 $\pm$(0.3)}&
88.4 $\pm$(0.6)&41.5 $\pm$(0.5)&87.4 $\pm$(1.2)\\

\hline 

Ours&\setlength{\fboxsep}{0pt}\colorbox{yellow}{98.6$\pm$(0.1)}&
\setlength{\fboxsep}{0pt}\colorbox{yellow}{80.4 $\pm$(0.5)}&
\setlength{\fboxsep}{0pt}\colorbox{yellow}{93.2 $\pm$(0.5)}&
\setlength{\fboxsep}{0pt}\colorbox{silver}{58.6 $\pm$(0.9)}&
\setlength{\fboxsep}{0pt}\colorbox{yellow}{86.9 $\pm$(1.3)}&
\setlength{\fboxsep}{0pt}\colorbox{yellow}{98.4 $\pm$(0.1)}&
\setlength{\fboxsep}{0pt}\colorbox{white}{79.4 $\pm$(0.7)}
&\setlength{\fboxsep}{0pt}\colorbox{yellow}{94.9 $\pm$(0.1)}&
\setlength{\fboxsep}{0pt}\colorbox{silver}{61.1 $\pm$(0.6)}&
\setlength{\fboxsep}{0pt}\colorbox{brown}{90.1  $\pm$(0.3)}
\\

\hline
\end{tabular}
\label{tab:supervised knn}
\end{center}
\end{table*}

\begin{table}[t]
\scriptsize
\caption{\textcolor{black}{This table includes the experimental results of the evaluation metric of Mutual Information Neural Estimation (MINE). Note that a higher score demonstrates a stronger dependency between the embedding feature and the corresponding labels (labels have been encoded into a one-hot vector).}}
\begin{center}

\setlength\tabcolsep{.4pt}
\begin{tabularx}{\textwidth}{l||c|c|c|c|c|c}
\cline{1-7}
\multirow{2}{*}{\shortstack[l]{\textbf{Method/} \\ \textbf{Dataset}}}& \multicolumn{6}{c}{\textbf{MINE}}  \\
\cline{2-7}
& \textbf{E-MNIST}&\textbf{F-MNIST}&\textbf{Reuters}&\textbf{20news}&\textbf{FSDD}&\textbf{MSI}\\ \cline{1-7}

PCA&1.88$\pm$(0.04)&1.67$\pm$(0.02)&1.00$\pm$(0.02)&1.39$\pm$(0.03)&1.79$\pm$(0.02)&\setlength{\fboxsep}{0.pt}\colorbox{silver}{2.18$\pm$(0.03)}\\

NMF&1.71$\pm$(0.01)&1.59$\pm$(0.02)&0.95$\pm$(0.01)&1.29$\pm$(0.02)&1.62$\pm$(0.04)&1.91$\pm$(0.09)\\
LLE&N/A&1.15$\pm$(0.08)&0.83$\pm$(0.02)&\setlength{\fboxsep}{0.pt}\colorbox{yellow}{1.69$\pm$(0.05)}&1.35$\pm$(0.05)&1.63$\pm$(0.04)\\
ISOMAP&N/A&1.62$\pm$(0.03)&0.85$\pm$(0.03)&1.18$\pm$(0.06)&1.62$\pm$(0.01)& 2.08$\pm$(0.06)\\

AE&\setlength{\fboxsep}{0.pt}\colorbox{brown}{2.15$\pm$(0.02)}&\setlength{\fboxsep}{0.pt}\colorbox{brown}{1.76$\pm$(0.02)}&\setlength{\fboxsep}{0.pt}\colorbox{brown}{1.00$\pm$(0.01)}&\setlength{\fboxsep}{0.pt}\colorbox{brown}{1.49$\pm$(0.03)}&\setlength{\fboxsep}{0.pt}\colorbox{brown}{1.76$\pm$(0.03)}&\setlength{\fboxsep}{0.pt}\colorbox{brown}{2.10$\pm$(0.07)}\\

AAE&2.11$\pm$(0.02)&1.74$\pm$(0.02)&0.99$\pm$(0.03)&\setlength{\fboxsep}{0.pt}\colorbox{white}{1.47$\pm$(0.02)}&1.67$\pm$(0.07)&1.70$\pm$(0.10)\\

SAE&2.08$\pm$(0.02)&1.60$\pm$(0.02)&0.97$\pm$(0.02)
&1.39$\pm$(0.02)&1.44$\pm$(0.06)&1.23$\pm$(0.08)\\

DAE&\setlength{\fboxsep}{0.pt}\colorbox{silver}{2.18$\pm$(0.02)}&\setlength{\fboxsep}{0.pt}\colorbox{silver}{1.77$\pm$(0.04)}&\setlength{\fboxsep}{0.pt}\colorbox{silver}{1.02$\pm$(0.02)}&\setlength{\fboxsep}{0.pt}\colorbox{brown}{1.49$\pm$(0.02)}&\setlength{\fboxsep}{0.pt}\colorbox{silver}{1.77$\pm$(0.06)}&\setlength{\fboxsep}{0.pt}\colorbox{silver}{2.18$\pm$(0.02)}\\

VAE&2.08$\pm$(0.03)&1.63$\pm$(0.03)&0.96$\pm$(0.03)
&1.45$\pm$(0.03)
&1.62$\pm$(0.07)&1.47$\pm$(0.07)\\

CPC&1.77$\pm$(0.03)&1.66$\pm$(0.05)&0.90$\pm$(0.01)&0.98$\pm$(0.01)&\setlength{\fboxsep}{0.pt}\colorbox{silver}{1.77$\pm$(0.04)}&1.89$\pm$(0.06)\\

\cline{1-7}

Ours&\setlength{\fboxsep}{0.pt}\colorbox{yellow}{2.22$\pm$(0.03)}&\setlength{\fboxsep}{0.pt}\colorbox{yellow}{1.81$\pm$(0.02)}&\setlength{\fboxsep}{0.pt}\colorbox{yellow}{1.05$\pm$(0.02)}&\setlength{\fboxsep}{0.pt}\colorbox{silver}{1.67$\pm$(0.05)}&\setlength{\fboxsep}{0.pt}\colorbox{yellow}{1.92$\pm$(0.05)}&\setlength{\fboxsep}{0pt}\colorbox{yellow}{2.20$\pm$(0.07)}\\

\cline{1-7}
\end{tabularx}
\label{tab:mi}
\end{center}
\end{table}

\begin{figure*}[t]
    \centering
    \includegraphics[width=0.8\textwidth]{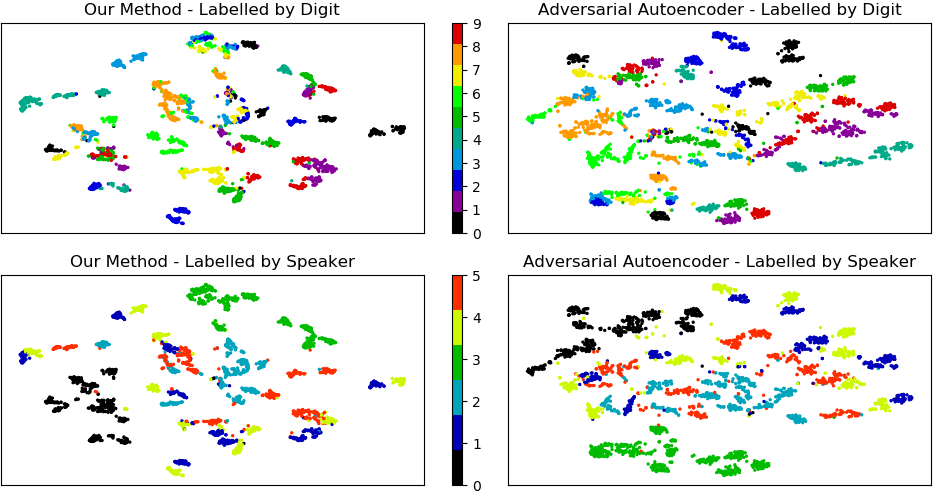}
    \caption{Four scatter plots of our proposed framework and the AAE on the FSDD dataset are shown in this diagram. The ground truth labels of the matching digits are used to colourize the top line charts. The bottom line shows the same scatters coloured by the labels of the speakers.}
    \label{fig:scatter}
\end{figure*}
\subsection{Parameters Settings}

In unsupervised learning, the selection of the hyper-parameters is usually chosen empirically based on pilot studies on the datasets where ground truth is known. Therefore, in order to make our proposed method competitive, we fixed all hyper-parameters to constant values among the experiments of the main setup, except for the scalar ($\epsilon$), which is proportional to the $\ell_2$-norm of each dataset. Specifically, for the Sinkhorn-Knopp algorithm \cite{NIPS2013_af21d0c9}, we find that $10$ iterations are sufficient to provide a reasonable solution. For the VAT regularisation term, the first hyper-parameter $\xi$ of random perturbations is set to $10$ across all datasets. We assign the second scaling hyper-parameter ($\epsilon$) of $r_{vadv}$ perturbations proportional to the average norm of each training set. We use a ratio of $\approx 1/5$ to the average $l_2\text{-norm}$ of the data-points in the dataset. This ratio is applied to \textcolor{black}{E}-MNIST ($\epsilon=2$), F-MNIST ($\epsilon=2$), FSDD ($\epsilon=5$), and MSI ($\epsilon=1$). Considering the \textit{tf-idf} measurement of word indexing in the text datasets and the high sparsity, the data-points are not normalised to the interval of $[0,1]$; instead, the highest value can reach $\sqrt{2,000}$. Hence, we set $\epsilon=20$ so that the VAT perturbations can weight the sparse data points in a more reasonable way. This value is applied to both text datasets. The settings for adapting the $\lambda$ values are kept constant as discussed in the Subsection \ref{section:preventing segmentation}. \par

We optimise all models for $5,000$ training epochs with a batch size of $256$. The Adam optimizer \cite{DBLP:journals/corr/KingmaB14} is utilised across all datasets and compared methods. We selected this optimization method as it demonstrates a fast convergence rate. The initial learning rate is selected to be $5\times10^{-4}$ with a drop to $1\times10^{-4}$ after the completion of $1,000$ training epochs.

\begin{table*}[t]
\caption{Comparative clustering results through \textit{k}-means for the extracted features of each method. The results of the best three methods are highlighted. Note that all methods, across all datasets, are evaluated for five independent runs and the average recorded performance with the corresponding deviations is reported. \textcolor{black}{Here, we report the standard metrics for evaluating the performance of the unsupervised methods, i.e., accuracy (ACC), the normalised mutual information (NMI), and the adjusted rand index (ARI).}}
\begin{center}
\setlength\tabcolsep{0.2pt}
\begin{tabularx}{\textwidth}{l| Y Y Y|Y Y Y|Y Y Y|Y Y Y|Y Y Y||X}
\hline

\multirow{3}{*}{\shortstack[l]{\textbf{Method/} \\ \textbf{Dataset}}}& \multicolumn{15}{c||}{\textbf{Unsupervised Clustering via \textit{k}-means Algorithm}}  & \multicolumn{1}{c}{\textbf{Linear Regression}}  \\
\cline{2-17}

& \multicolumn{3}{c}{\textbf{E-MNIST}}  & \multicolumn{3}{c}{\textbf{F-MNIST}}  & \multicolumn{3}{c}{\textbf{Reuters}} & \multicolumn{3}{c}{\textbf{20news}} & \multicolumn{3}{c||}{\textbf{FSDD}} & \multicolumn{1}{c}{\textbf{MSI}}\\

\cline{2-17}
& \multicolumn{1}{c}{ACC} & \multicolumn{1}{c}{NMI} & \multicolumn{1}{c}{ARI} &  \multicolumn{1}{c}{ACC} & \multicolumn{1}{c}{NMI} & \multicolumn{1}{c}{ARI} &  \multicolumn{1}{c}{ACC} & \multicolumn{1}{c}{NMI} & \multicolumn{1}{c}{ARI} &  \multicolumn{1}{c}{ACC} & \multicolumn{1}{c}{NMI} & \multicolumn{1}{c}{ARI} &  \multicolumn{1}{c}{ACC} & \multicolumn{1}{c}{NMI} & \multicolumn{1}{c||}{ARI} & \multicolumn{1}{c}{RSS} \\ \hline

PCA&52.8 $\pm$(0.1)&45.7 $\pm$(0.1) &34.5 $\pm$(0.3)&49.0 $\pm$(0.8)&50.9 $\pm$(0.6)&34.9 $\pm$(0.0)&53.9 $\pm$(0.1)&40.6 $\pm$(0.2)&27.5 $\pm$(0.1)&29.4 $\pm$(0.8)&32.8 $\pm$(0.2)&13.5 $\pm$(0.3)&29.4 $\pm$(1.2)&18.9 $\pm$(1.9)&10.4 $\pm$(1.3)&\multicolumn{1}{c}{2073.0 $\pm$(3.9)}\\

NMF&48.1 $\pm$(0.4)&39.5 $\pm$(0.3)&28.0 $\pm$(0.2)&53.9 $\pm$(0.1)&52.7 $\pm$(0.8)&36.4 $\pm$(0.8)&58.6 $\pm$(1.0)&41.0 $\pm$(0.2)&28.1 $\pm$(1.1)&28.8 $\pm$(2.0)&32.1 $\pm$(0.3)&12.2 $\pm$(0.3)&34.5 $\pm$(0.6)&30.0 $\pm$(0.2)&17.2 $\pm$(0.5)&\multicolumn{1}{c}{3016.9 $\pm$(23.6)}\\

LLE&N/A&N/A&N/A&37.2 $\pm$(1.4)&33.2 $\pm$(2.3)&19.7 $\pm$(1.8)&44.0 $\pm$(3.2)&13.2 $\pm$(1.4)&2.3 $\pm$(0.9)&\setlength{\fboxsep}{0pt}\colorbox{yellow}{49.2} \setlength{\fboxsep}{0pt}\colorbox{yellow}{$\pm$(1.1)}&\setlength{\fboxsep}{0pt}\colorbox{yellow}{50.0} \setlength{\fboxsep}{0pt}\colorbox{yellow}{$\pm$(1.1)}&\setlength{\fboxsep}{0pt}\colorbox{yellow}{33.2} \setlength{\fboxsep}{0pt}\colorbox{yellow}{$\pm$(2.1)}&35.2 $\pm$(0.2)&28.7 $\pm$(0.1)&13.2 $\pm$(0.2)&\multicolumn{1}{c}{\setlength{\fboxsep}{0pt}\colorbox{brown}{1859.2 $\pm$(41.5)}}\\

ISOMAP&N/A&N/A&N/A&56.8 $\pm$(1.3)&57.1 $\pm$(1.0)&42.1 $\pm$(1.5)&
\setlength{\fboxsep}{0pt}\colorbox{yellow}{73.0} \setlength{\fboxsep}{0pt}\colorbox{yellow}{$\pm$(0.1)}&\setlength{\fboxsep}{0pt}\colorbox{silver}{52.5} \setlength{\fboxsep}{0pt}\colorbox{silver}{$\pm$(0.1)}&\setlength{\fboxsep}{0pt}\colorbox{yellow}{58.5} \setlength{\fboxsep}{0pt}\colorbox{yellow}{$\pm$(0.0)}&39.1 $\pm$(0.6)&40.5 $\pm$(0.3)&22.2 $\pm$(0.2)&37.1 $\pm$(0.4)&33.1 $\pm$(0.7)&18.2 $\pm$(0.2)&\multicolumn{1}{c}{\setlength{\fboxsep}{0pt}\colorbox{white}{1885.3 $\pm$(4.6)}}
\\

AE&86.2 $\pm$(1.5)&81.6 $\pm$(1.1)&77.2 $\pm$(1.8)&\setlength{\fboxsep}{0pt}\colorbox{silver}{58.2} \setlength{\fboxsep}{0pt}\colorbox{silver}{$\pm$(1.7)}&\setlength{\fboxsep}{0pt}\colorbox{silver}{64.4} \setlength{\fboxsep}{0pt}\colorbox{silver}{$\pm$(1.5)}&\setlength{\fboxsep}{0pt}\colorbox{silver}{46.6} \setlength{\fboxsep}{0pt}\colorbox{silver}{$\pm$(0.5)}&58.6 $\pm$(3.7)&38.0 $\pm$(5.8)&32.9 $\pm$(7.5)&\setlength{\fboxsep}{0pt}\colorbox{brown}{47.0} \setlength{\fboxsep}{0pt}\colorbox{brown}{$\pm$(0.9)}&\setlength{\fboxsep}{0pt}\colorbox{brown}{41.0} \setlength{\fboxsep}{0pt}\colorbox{brown}{$\pm$(1.0)}&\setlength{\fboxsep}{0pt}\colorbox{brown}{29.2} \setlength{\fboxsep}{0pt}\colorbox{brown}{$\pm$(0.8)}&\setlength{\fboxsep}{0pt}\colorbox{silver}{38.8} \setlength{\fboxsep}{0pt}\colorbox{silver}{$\pm$(0.9)}&\setlength{\fboxsep}{0pt}\colorbox{silver}{34.0} \setlength{\fboxsep}{0pt}\colorbox{silver}{$\pm$(1.0)}&\setlength{\fboxsep}{0pt}\colorbox{silver}{20.1} \setlength{\fboxsep}{0pt}\colorbox{silver}{$\pm$(1.1)}&\multicolumn{1}{c}{\setlength{\fboxsep}{0pt}\colorbox{white}{1885.8 $\pm$(46.9)}}
\\

AAE&59.9 $\pm$(2.4)&46.9 $\pm$(1.7)&37.9 $\pm$(2.5)&42.8 $\pm$(1.5)&35.5  $\pm$(1.7)&24.1 $\pm$(1.7)&\setlength{\fboxsep}{0pt}\colorbox{brown}{69.2} \setlength{\fboxsep}{0pt}\colorbox{brown}{$\pm$(1.0)}&\setlength{\fboxsep}{0pt}\colorbox{brown}{47.4} \setlength{\fboxsep}{0pt}\colorbox{brown}{$\pm$(1.4)}&\setlength{\fboxsep}{0pt}\colorbox{brown}{46.7} \setlength{\fboxsep}{0pt}\colorbox{brown}{$\pm$(1.0)}&43.2 $\pm$(2.1)&35.4 $\pm$(2.0)&24.5 $\pm$(1.9)&32.2 $\pm$(3.1)&26.3 $\pm$(4.6)&15.2 $\pm$(3.0)&\multicolumn{1}{c}{2653.1 $\pm$(257.9)}
\\

SAE&\setlength{\fboxsep}{0pt}\colorbox{silver}{91.0} \setlength{\fboxsep}{0pt}\colorbox{silver}{$\pm$(4.4)}&\setlength{\fboxsep}{0pt}\colorbox{silver}{84.6} \setlength{\fboxsep}{0pt}\colorbox{silver}{$\pm$(3.1)}&\setlength{\fboxsep}{0pt}\colorbox{silver}{83.2} \setlength{\fboxsep}{0pt}\colorbox{silver}{$\pm$(5.2)}&\setlength{\fboxsep}{0pt}\colorbox{brown}{57.7} \setlength{\fboxsep}{0pt}\colorbox{brown}{$\pm$(3.9)}&58.8 $\pm$(2.6)&42.9 $\pm$(4.0)&46.9 $\pm$(5.6)&14.9 $\pm$(5.8)&12.0 $\pm$(5.1)&33.9 $\pm$(1.8)&32.3 $\pm$(0.6)&16.2 $\pm$(0.5)&30.7 $\pm$(1.3)&23.1 $\pm$(1.0)&11.6 $\pm$(0.8)&\multicolumn{1}{c}{2015.4 $\pm$(82.5)}
\\

DAE&\setlength{\fboxsep}{0pt}\colorbox{brown}{87.0} \setlength{\fboxsep}{0pt}\colorbox{brown}{$\pm$(2.3)}&
\setlength{\fboxsep}{0pt}\colorbox{brown}{82.0} \setlength{\fboxsep}{0pt}\colorbox{brown}{$\pm$(1.6)}&\setlength{\fboxsep}{0pt}\colorbox{brown}{77.9} \setlength{\fboxsep}{0pt}\colorbox{brown}{$\pm$(2.8)}&57.3 $\pm$(3.1)&\setlength{\fboxsep}{0pt}\colorbox{brown}{63.8} \setlength{\fboxsep}{0pt}\colorbox{brown}{$\pm$(2.0)}&\setlength{\fboxsep}{0pt}\colorbox{brown}{45.7} \setlength{\fboxsep}{0pt}\colorbox{brown}{$\pm$(1.8)}&64.4 $\pm$(5.1)&39.0 $\pm$(6.0)&37.2 $\pm$(12.2)&\setlength{\fboxsep}{0pt}\colorbox{white}{45.4} \setlength{\fboxsep}{0pt}\colorbox{white}{$\pm$(1.5)}&40.0 $\pm$(0.8)&26.1 $\pm$(0.9)&\setlength{\fboxsep}{0pt}\colorbox{brown}{37.4} \setlength{\fboxsep}{0pt}\colorbox{brown}{$\pm$(0.3)}&\setlength{\fboxsep}{0pt}\colorbox{brown}{32.3} \setlength{\fboxsep}{0pt}\colorbox{brown}{$\pm$(0.7)}&\setlength{\fboxsep}{0pt}\colorbox{brown}{18.6} \setlength{\fboxsep}{0pt}\colorbox{brown}{$\pm$(0.5)}&\multicolumn{1}{c}{1900.5$\pm$(73.1)}
\\

VAE&74.9 $\pm$(6.1)&65.8 $\pm$(3.7)&59.8 $\pm$(5.4)&48.9 $\pm$(2.5)&45.4 $\pm$(2.9)&31.1 $\pm$(2.2)&67.6 $\pm$(1.4)&46.6 $\pm$(2.0)&\setlength{\fboxsep}{0pt}\colorbox{white}{45.4} \setlength{\fboxsep}{0pt}\colorbox{white}{$\pm$(1.7)}&\setlength{\fboxsep}{0pt}\colorbox{white}{47.0} \setlength{\fboxsep}{0pt}\colorbox{white}{$\pm$(1.4)}&\setlength{\fboxsep}{0pt}\colorbox{white}{40.7} \setlength{\fboxsep}{0pt}\colorbox{white}{$\pm$(0.3)}&\setlength{\fboxsep}{0pt}\colorbox{white}{28.9} \setlength{\fboxsep}{0pt}\colorbox{white}{$\pm$(0.7)}&35.6 $\pm$(2.7)&28.2 $\pm$(2.5)&16.6 $\pm$(2.3)&\multicolumn{1}{c}{2475.4 $\pm$(105.9)}
\\

CPC &34.0 $\pm$(0.6)&28.9 $\pm$(0.7)&15.2 $\pm$(0.8)&41.9 $\pm$(1.7)&38.4 $\pm$(0.7)&23.7 $\pm$(1.2)&47.8 $\pm$(1.1)&10.5 $\pm$(0.5)&10.3 $\pm$(0.6)&22.2 $\pm$(0.8)&25.4 $\pm$(0.6)&9.9 $\pm$(0.3)&36.5 $\pm$(1.8)&28.8 $\pm$(2.7)&16.6 $\pm$(1.8)&\multicolumn{1}{c}{\setlength{\fboxsep}{0pt}\colorbox{silver}{1793.7 $\pm$(24.6)}} \\

\hline 

Ours&\setlength{\fboxsep}{0pt}\colorbox{yellow}{96.4} \setlength{\fboxsep}{0pt}\colorbox{yellow}{$\pm$(0.6)}&
\setlength{\fboxsep}{0pt}\colorbox{yellow}{92.0} \setlength{\fboxsep}{0pt}\colorbox{yellow}{$\pm$(0.9)}&
\setlength{\fboxsep}{0pt}\colorbox{yellow}{92.3} \setlength{\fboxsep}{0pt}\colorbox{yellow}{$\pm$(1.3)}&
\setlength{\fboxsep}{0pt}\colorbox{yellow}{65.0} \setlength{\fboxsep}{0pt}\colorbox{yellow}{$\pm$(1.8)}&
\setlength{\fboxsep}{0pt}\colorbox{yellow}{67.0} \setlength{\fboxsep}{0pt}\colorbox{yellow}{$\pm$(1.1)}&
\setlength{\fboxsep}{0pt}\colorbox{yellow}{51.9} \setlength{\fboxsep}{0pt}\colorbox{yellow}{$\pm$(2.1)}&
\setlength{\fboxsep}{0pt}\colorbox{silver}{70.7} \setlength{\fboxsep}{0pt}\colorbox{silver}{$\pm$(3.6)}&
\setlength{\fboxsep}{0pt}\colorbox{yellow}{60.1} \setlength{\fboxsep}{0pt}\colorbox{yellow}{$\pm$(4.3)}&
\setlength{\fboxsep}{0pt}\colorbox{silver}{54.4} \setlength{\fboxsep}{0pt}\colorbox{silver}{$\pm$(2.9)}&
\setlength{\fboxsep}{0pt}\colorbox{silver}{47.8} \setlength{\fboxsep}{0pt}\colorbox{silver}{$\pm$(1.7)}&
\setlength{\fboxsep}{0pt}\colorbox{silver}{49.0} \setlength{\fboxsep}{0pt}\colorbox{silver}{$\pm$(0.8)}&
\setlength{\fboxsep}{0pt}\colorbox{silver}{31.4} \setlength{\fboxsep}{0pt}\colorbox{silver}{$\pm$(2.2)}&
\setlength{\fboxsep}{0pt}\colorbox{yellow}{39.1} \setlength{\fboxsep}{0pt}\colorbox{yellow}{$\pm$(1.5)}&
\setlength{\fboxsep}{0pt}\colorbox{yellow}{36.2} \setlength{\fboxsep}{0pt}\colorbox{yellow}{$\pm$(1.9)}&
\setlength{\fboxsep}{0pt}\colorbox{yellow}{20.0} \setlength{\fboxsep}{0pt}\colorbox{yellow}{$\pm$(1.0)}&\multicolumn{1}{c}{\setlength{\fboxsep}{0pt}\colorbox{yellow}{1539.8 $\pm$(74.8)}}
\\

\hline
\end{tabularx}
\label{tab:k-means results}
\end{center}
\end{table*}

\subsection{Comparative Results}

Table \ref{tab:supervised knn} includes the experimental results of the two evaluation strategies. The header column with the title "Linear Classifier'' refers to the \textcolor{black}{accuracy of the} linear evaluation metric, where a single classification layer is trained, in a supervised manner, with the extracted features as the input and the ground truth (labels) as the target. The right side of Table \ref{tab:supervised knn}, with the column name "Graph Evaluation'', presents the results of the evaluation strategy based on the t-SNE graph. We mark a positive prediction if and only if the majority of the labels of the nearest points match the relevant point. In both evaluation strategies, the ground truth is used for evaluation only and should never be used in training the encoder. For convenience, we highlighted the top three methods according to the order of the following colours: yellow, gray, and bronze. All recorded results are computed via five independent runs for each method. The mean accuracy and the standard deviation are reported, respectively. \textcolor{black}{Note that the original implementation of the CPC study applies transformation functions specifically to visual datasets, while in our work we avoid explicitly transformation functions designated for any particular types of datasets. Similarly, the text dataset under comparison in the CPC study is preprocessed with the Word2Vec \cite{mikolov2013efficient} feature extraction approach. However, to  fairly evaluate all the methods under the validation process, in this work, we consider the same input across all the models (including CPC) as those discussed in Section \ref{section:experimental section} without adopting additional external preprocessing methodologies or transformation functions as those discussed in CPC.}\par

Overall, our proposed method outperforms all compared methods according to the linear classifier\textcolor{black}{, except in the 20news dataset where LLE performs slightly better. We observe a reasonable margin, especially on datasets of FSDD and Reuters, while a lower margin is captured in E-MNIST, 20news and F-MNIST.} This implies that our method locates data points with similar semantics in a close spatial neighbourhood that is linearly separable at the latent representation. This can be attributed to the VAT perturbations since the training process enables the model to be invariant to negligible transformations. Additionally, despite the different dimensions of the different data types and the variations of data points, the effectiveness of the proposed framework is clearly demonstrated in terms of the broadly used linear evaluation approach. \par

Regarding the t-SNE graph evaluation (right side of Table \ref{tab:supervised knn}), our proposed method achieves its best results on two of the datasets, while producing competitive results on 20news and FSDD. Despite the encouraging results according to the linear evaluation, our method is not included in the top three algorithms on F-MNIST. Note that there is no well-accepted method to precisely evaluate a visualisation method. Nevertheless, we visualise the generated graphs of the best recorded method and our method for the FSDD dataset to better understand the distribution in the scatter plots. For FSDD, \textcolor{black}{we can analyse} the visualisation results from two perspectives: 1) the ground truth labels, where each recording audio represents a digit, and 2) as recordings from different speakers. These two observations are visualised in Fig. \ref{fig:scatter}, where in the top row the data points are colourized based on their labels and the bottom row based on the speakers' identity. The left column of Fig.  \ref{fig:scatter} represents our method, while the right column represents AAE. Through these plots, we observe that, in certain circumstances, both methods mix the points. For instance, all recordings of speakers '3' (green colour, bottom row) and '0' (black colour, bottom row) are located in the same neighbourhood from both embedding techniques but not for the remaining speakers. By contrast, the audios of digit '4' (middle green, top row) are placed in the same spatial location for four speakers in both algorithms. As a result, we conclude that, aside from the ground truth, which can be viewed as an evaluation metric, the spatial location of the points can also be driven by different patterns, and we anticipate that this will be the case for F-MNIST, where our algorithm achieves a $1.3\%$ margin in linear evaluation. \par

\textcolor{black}{The estimation of mutual information based on the MINE \cite{pmlr-v80-belghazi18a} validation methodology is presented in Table \ref{tab:mi}. By analysing the results in the table, we observe that our proposed model demonstrates the highest dependency between the bottleneck representation and the ground truth representation (one-hot vector in the case of labels) except for the 20news dataset. However, LLE achieves the highest accuracy in linear evaluation, so this achievement is not unexpected here. This observation is for all the datasets under the benchmark and regardless of the architecture of the model.} \par

Table \ref{tab:k-means results} includes the results of the \textit{k}-means method (middle head column). By examining the results, we find our proposed model achieves the highest performance on the majority of the datasets and metrics under comparison, except that Isomap surpasses in the two unsupervised metrics on the Reuters dataset \textcolor{black}{and LLE on the 20news dataset}. The selected subset of Reuters \textcolor{black}{and the 20news dataset are highly unbalanced in population. Our method aims to uniformly spread into latent representation, and perhaps this might be the reason that Isomap and LLE achieve slightly better performance in text datasets}. Nonetheless, we notice that on specific datasets such as F-MNIST, our method outperforms  the best competitive algorithm by $\approx 3\%$, despite the fact that in graph evaluation the performance was lower. \par

The last single column of Table \ref{tab:k-means results} includes the results of the evaluation strategy of linear regression. Note that the lower result indicates the better performance. Here, we observe that our model outperforms all the methods in the comparison list by a large margin. This can be ascribed to the training process of our method, which is invariant to small perturbations (VAT). Therefore, our proposed model demonstrates robustness by identifying hidden patterns across the spectras and ignoring possible interference from surrounding chemical components on the corresponding spectra. \par

\begin{figure}[t]
    \centering
    \includegraphics[width=0.5 \textwidth]{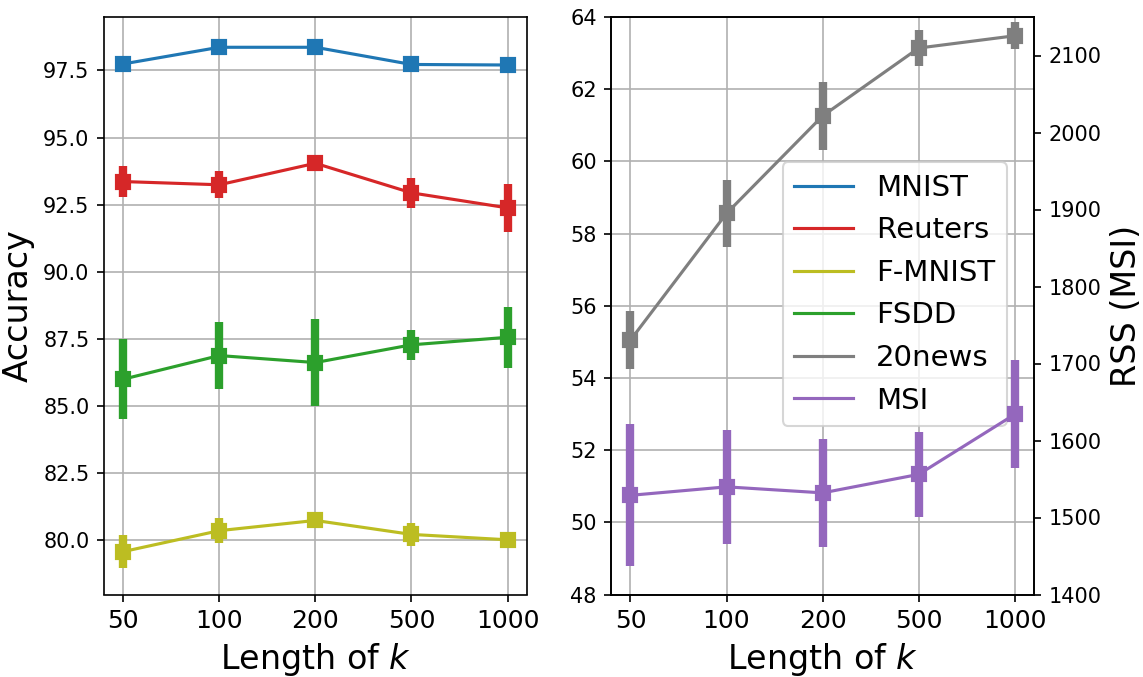}
    \caption{This figure presents the impact of the output dimensions $k$ ($x$-axis) in terms of 1) the accuracy left ($y$-axis) of the model in linear classification evaluation among the comparison dataset and 2) the RSS (right ($y$-axis)) of the linear regression for MSI data only.}
   \label{fig:plot k}
\end{figure}

\begin{figure}[t]
    \centering
    \includegraphics[width=0.5 \textwidth]{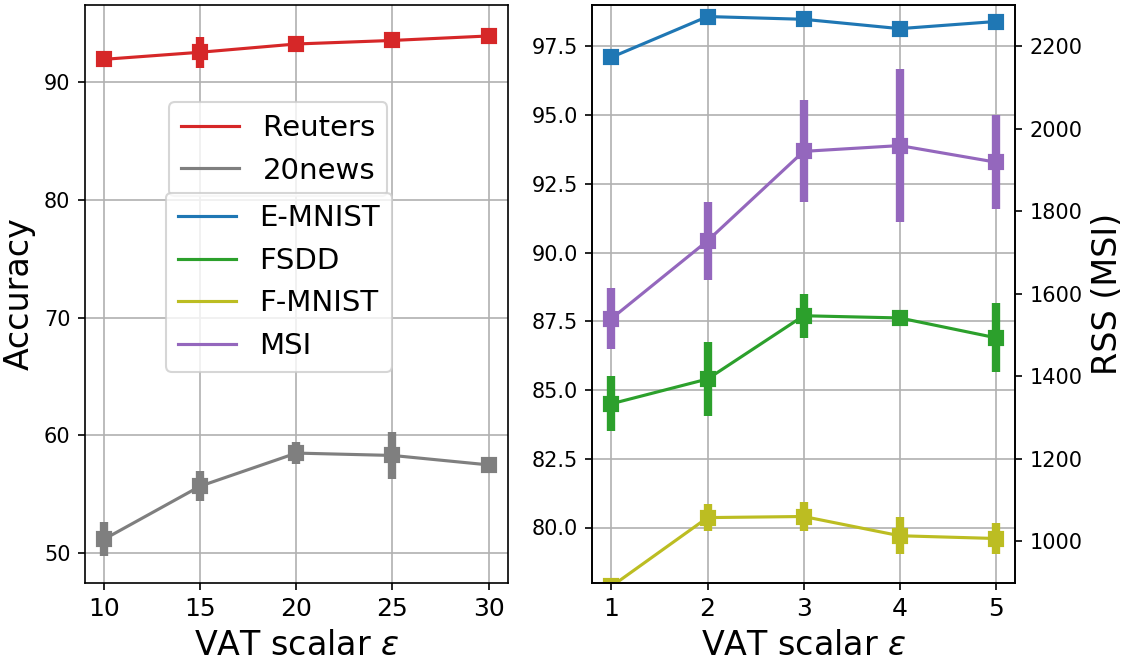}
    \caption{The influence of the hyper-parameter $\epsilon$ on the framework's accuracy. $x$-axis corresponds to the value of the scalar. Left $y$-axis represents the linear evaluation of the classifier, while the right $y$-axis refers to the linear regression (MSI only data).}
   \label{fig:plot vat_e}
\end{figure}

\subsection{Impact of $k$}

The effect of dimension $k$, which refers to the output layer of module $g_\psi$, is investigated in this subsection. As for the main setup, all the remaining hyper-parameters are kept constant. We vary $k$ in the interval of $[50, 1000]$ with a sensible step. We use the linear evaluation strategy as the most direct and non-stochastic metric and plot the average accuracy and standard deviation over five independent runs. Our findings are illustrated in Fig. \ref{fig:plot k}, where we see \textcolor{black}{an improvement} on F-MNIST and 20news as $k$ increases. This may indicate that a larger dimension in the output layer will have the potential to capture more hidden patterns, in a classification task, resulting in \textcolor{black}{a} slightly higher accuracy on most challenging datasets. By contrast, the evaluation of linear regression of MSI input data (right $y$-axis of the right plot in Fig. \ref{fig:plot k}) demonstrates a slightly higher sensitivity to $k$, where the highest difference is observed at $k=1,000$\textcolor{black}{,} with the average performance increased by $\approx 80$ points with the main setup.

\subsection{Experiments on $\epsilon$ in VAT}

For this setup, we set all hyper-parameters based on the main experimental settings and only modify the scalar value of $\epsilon$ of the VAT process. We alter the interval of $\epsilon = [1,5]$ for the datasets of \textcolor{black}{E}-MNIST, F\textcolor{black}{-}MNIST, FSDD\textcolor{black}{,} and MSI, whose values are normalised to the range of $[0,1]$. On the other hand, we define the range between $[10,30]$ for the text datasets (whose highest value is $44$) with a step-size of five. The quantity measurements are plotted in Fig. \ref{fig:plot vat_e}, where the $x$-axis refers to the value assigned to the scalar ($\epsilon$). Similar to the previous experiment, we report the accuracy of the linear classifier (\textcolor{black}{left panel}) and the linear regression (\textcolor{black}{right panel}) on the MSI data. By examining the plots, we notice a low fluctuation in accuracy on all normalised datasets (refer to the right side of Fig. \ref{fig:plot vat_e}), despite the different values of $\epsilon$. Similar results are also observed on the Reuters text set. The performance of our proposed model degrades on 20news when $\epsilon=10$. A possible explanation for this degradation is that the low weight value of VAT perturbations relative to the values of the data points fails to provide enough influence on the model to learn invariant representations. Nonetheless, we find that for higher values of $\epsilon \geq 15$ , the performance is competitive and the performance remains stable. According to the linear regression evaluation, our model is less accurate in comparison to the main setting. This is expected as the $\ell_2$-norm of the dataset is low ($\approx 5$). Hence by increasing the VAT scalar ($2 \leq \epsilon$) the input points are transformed significantly.

\begin{table}[t]
\caption{Evaluation of different transformation operations. MSI is evaluated in RSS metric only (lower the better).}
\setlength\tabcolsep{1.5pt}
\begin{tabularx}{250.pt}{l|c|c|c|c|c|c}
\hline
\multirow{2}{*}{\begin{tabular}{@{}c@{}}\end{tabular}}& \multicolumn{1}{c}{\textbf{E-MNIST}}  & \multicolumn{1}{c}{\textbf{F-MNIST}}  & \multicolumn{1}{c}{\textbf{Reuters}} & \multicolumn{1}{c}{\textbf{20news}} & \multicolumn{1}{c}{\textbf{FSDD}}& \multicolumn{1}{c}{\textbf{MSI}}  \\
\hline
NP&32.8$\pm$(5.8)&54.2$\pm$(3.4)&76.6$\pm$(3.6)&19.0$\pm$(2.8)&37.9$\pm$(5.8)& \begin{tabular}{@{}c@{}}2277.0\\$\pm$(172.1)\end{tabular}\\
RP&94.9$\pm$(0.5)&73.1$\pm$(0.9)&91.5$\pm$(1.0)&52.1$\pm$(1.1)&53.1$\pm$(4.9)& \begin{tabular}{@{}c@{}}2226.2\\$\pm$(83.8)\end{tabular}\\

\begin{tabular}{@{}c@{}}RP+\\VAT\end{tabular}&98.3$\pm$(0.5)&79.7$\pm$(0.9)&94.5$\pm$(0.1)&58.2$\pm$(1.3)&84.7$\pm$(3.7)& \begin{tabular}{@{}c@{}}1576.6\\$\pm$(120.9)\end{tabular}\\
VAT &98.6$\pm$(0.1)& 80.4$\pm$(0.3)& 93.2$\pm$(0.5)& 58.6$\pm$(0.9)& 86.9$\pm$(1.3)& \begin{tabular}{@{}c@{}}1539.8\\$\pm$(74.8)\end{tabular}\\
\hline
\end{tabularx}
\label{tab: effect of vat}
\end{table}

\subsection{Ablation studies}

We investigate the contribution of each main component of the proposed framework by modifying the learning process in three aspects: 1) assessing the transformation impact of VAT perturbations and combining them with random perturbations; 2) evaluating the effectiveness of the batch normalisation layer; and 3) examining the adaptation policy of hyper-parameter $\lambda$ . We keep the main settings of the experiments across datasets and only vary the related module/parameters. The results presented in this subsection are averaged over five independent runs for each experiment, and the mean accuracy and the standard deviation are reported. Note that when we refer to the metric accuracy, we mean the performance of the linear classifier trained on top of the embedding space.

\subsubsection{Impact of transformations} Here we study the importance of VAT in the learning process of our framework. We include three additional experiments in \textcolor{black}{these} comparisons: 1) no perturbation (NP), the framework is optimised on raw data without transformation operations (e.g., VAT); 2) random perturbations (RP) are used exclusively to completely replace VAT perturbations; 3) a combination of random perturbations in the first transformation, and the second transformation is achieved through the VAT process (RP + VAT). The main setup is denoted with the initials VAT in the comparison. \par

\begin{table}[t]
\caption{Learning performance by not including the batch normalisation layer. Metric of MSI is on RSS measurement.}
\setlength\tabcolsep{1.5pt}
\begin{tabularx}{250.pt}{l|c|c|c|c|c|c}
\hline
\multirow{2}{*}{\begin{tabular}{@{}c@{}}\end{tabular}}& \multicolumn{1}{c}{\textbf{E-MNIST}}  & \multicolumn{1}{c}{\textbf{F-MNIST}}  & \multicolumn{1}{c}{\textbf{Reuters}} & \multicolumn{1}{c}{\textbf{20news}} & \multicolumn{1}{c}{\textbf{FSDD}} & \multicolumn{1}{c}{\textbf{MSI}} \\
\hline

No-BN&95.8$\pm$(0.5)& 77.3$\pm$(0.8)&93.0$\pm$(0.7)& 54.3$\pm$(1.1)& 82.4$\pm$(0.9)& \begin{tabular}{@{}c@{}}1666.0\\$\pm$(57.7)\end{tabular}\\
EN-BN&97.4$\pm$(0.2)&77.2$\pm$(0.2)&93.7$\pm$(0.4)&59.8$\pm$(1.0)&87.1$\pm$(1.0)& \begin{tabular}{@{}c@{}}1580.2\\$\pm$(115.3)\end{tabular}\\
CL-BN&97.3$\pm$(0.3)&78.5$\pm$(0.5)&93.4$\pm$(0.8)&56.2$\pm$(1.6)&75.5$\pm$(2.5)& \begin{tabular}{@{}c@{}}1913.4\\$\pm$(183.7)\end{tabular}\\
\begin{tabular}{@{}c@{}}EN-CL\\BN\end{tabular}&98.6$\pm$(0.1)& 80.4$\pm$(0.5)& 93.2$\pm$(0.5)& 58.6$\pm$(0.9)& 86.9$\pm$(1.3)& \begin{tabular}{@{}c@{}}1539.8\\$\pm$(74.8)\end{tabular}\\
\hline
\end{tabularx}
\label{tab: bn impact}
\end{table}

Table \ref{tab: effect of vat} presents the accuracy of the transformation process on the raw data. The results clearly show the model has failed to learn properly when no transformation function is used prior to (denoted as NP). The experimental results with random perturbations (RP) indicate an improvement, but not significant. On the other hand, the model enhances its performance with a combination of RP and VAT. However, both transformations obtained through the VAT method achieve the highest accuracy for linear classification and the lowest RSS for linear regression, confirming our decision to select the VAT for enforcing invariant representations.

\begin{figure}[t]
    \centering
    \includegraphics[width=0.5 \textwidth]{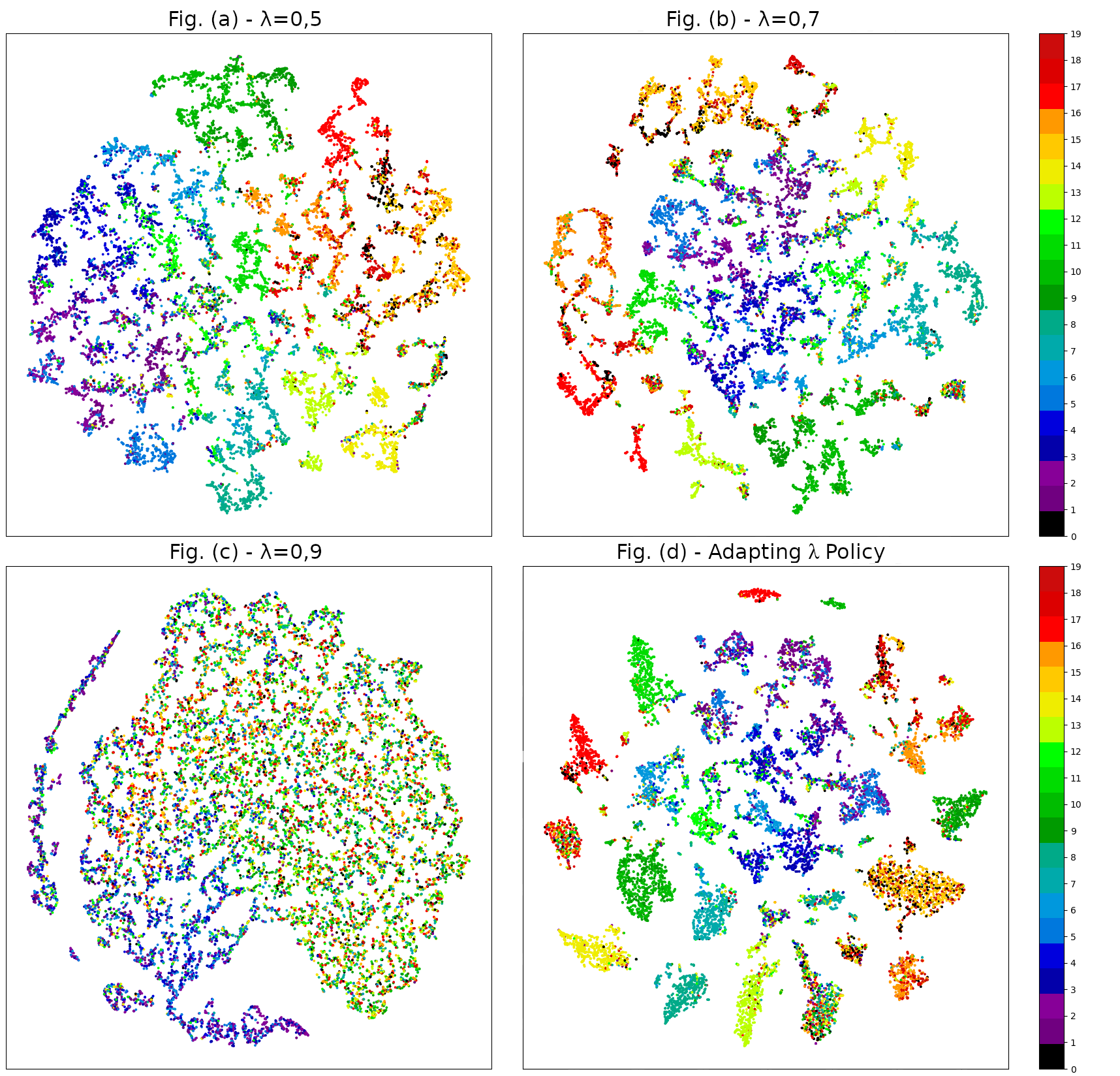}
    \caption{The influence of the hyper-parameter $\epsilon$ on the framework's accuracy. \textcolor{black}{All experiments are conducted in the 20news dataset.} $x$-axis denotes the value of the scalar.}
   \label{fig:plots}
\end{figure}
\subsubsection{Learning performance without batch normalisation (BN)} We modify the architecture of the main setup in three different variations: 1) no BN layers in both modules (No-BN); 2) only BN layers on the encoder module (EN-BN); 3) only BN layers on the classifier module (CL-BN). All instances of the model are optimised for $5,000$ epochs, and we report the results of the linear classifier. The results from the main setting \textcolor{black}{are} indicated with "EN-CL BN'' in the corresponding table.

Table \ref{tab: bn impact} shows that using BN in the encoder module enhances learning performance across all datasets. This also holds true if BN is applied in both modules\textcolor{black}{,} similar to the main setup. On the other hand, including BN only in the classification module does not provide any benefit to the framework and on specific datasets, i.e., 20news and FSDD, the performance is worse than with the No-BN structure.

\subsubsection{Adaptation policy of $\lambda$} We explore the \textcolor{black}{benefits} of our proposed adaptation policy in the model's training process. More specifically, we compare results when assigning different constant values to hyper-parameter $\lambda$. Unlike the previous results, in this subsection we trained the framework for a total of $10,000$ epochs across all experiments. All the remaining hyper-parameters use the settings for the main experiment. To demonstrate the distribution of the data in  the latent representation, we employ the visualisation method of t-SNE and project all plots \textcolor{black}{onto} a $2$-dimensional space. Here we present the results on 20news, and the corresponding classes are colourised accordingly. Figure \ref{fig:plots} includes the four plots, where $\lambda$ is set to 0.5 (Fig. \ref{fig:plots}.(a)), 0.7 (Fig. \ref{fig:plots}.(b)), 0.9 (Fig. \ref{fig:plots}.(c)), and tuned according to the adaptation policy (Fig. \ref{fig:plots}.(d)).  \par

From these results, we see that overall, the data points that hold a common semantic are grouped in the same spatial neighborhood. When $\lambda=0.5$ and $\lambda=0.7$ (the upper row in the figure), we observe that there are sub-classes within the same class, e.g., the green dots in the graphs. These reveal that the model sub-groups the data points in the embedding space. When $\lambda=0.9$, the model does not produce a meaningful representation as the sparsity of the generated target distribution is too low. On the other hand, we can observe from the results in \textcolor{black}{the} bottom right that through the proposed adapting policy, the model identifies groups with common patterns and locates them properly in the embedding space.

\section{Conclusion}

In this work, we present a generic framework for representation learning in a self-supervised methodology. Our main motivation is to design an invariant representation learning method that is independent of any specific data type or specific model architecture. This is achieved by introducing and modifying the virtual adversarial training process to form a contrastive objective with the aim of replacing the prior transformations. Additionally, we introduce a policy to adapt the smoothness of the target distribution in the Sinkhorn-Knopp algorithm. Extensive comparative and ablation studies verify that the proposed method can be applied to various data types and shows competitive performance robust to the main hyper-parameters in the proposed algorithm. \par

Despite the promising results exhibited by the proposed model, we observed a sensitivity of the performance to minor or aggressive transformations proportional to the input data in certain circumstances. This sensitivity is caused by assigning either a very low value to the defined weight scalar of the adversarial training process, which results in negligible augmentation of the data point, or a very high value, which leads to aggressive augmentation of the corresponding data point. Furthermore, a sub-grouping effect in the embedding space is also examined, which is related to the discrete output of the classification module. The proposed model successfully avoids sub-grouping the data points by implementing an adaptation policy for constraining the sparsity of target distributions at a maintained objective level. Therefore, our future work will first explore and develop more powerful self-supervised learning mechanisms for imposing invariant representations that do not rely on hyper-parameters, e.g., weighting scalars. Our second objective will consider inferring the latent representation in specific probabilistic behaviours, enabling us to select a constant hyper-parameter in the optimal transport and avoid the sub-grouping side effect.

{\small
\bibliographystyle{unsrt}  

\bibliography{ref}
}

\end{document}